%% file: main.tex
\documentclass[10pt,journal,compsoc]{IEEEtran}
\usepackage{amsmath,amsfonts}
\usepackage{algorithm}
\usepackage{algorithmic}
\usepackage{array}

\usepackage[caption=false,font=normalsize,labelfont=sf,textfont=sf]{subfig}
\usepackage{textcomp}
\usepackage{stfloats}
\usepackage{url}
\usepackage{verbatim}
\usepackage{graphicx}
\usepackage{booktabs}
\usepackage{multirow}
\usepackage{colortbl}
\usepackage{makecell}
\usepackage[most]{tcolorbox}
\newcommand{\et}[2]{$#1^{\pm #2}$}
\newcommand{\etb}[2]{$\mathbf{#1}^{\pm #2}$}
\newcommand{\ets}[2]{$\underline{#1}^{\pm #2}$}
\usepackage[table]{xcolor}

\def\BibTeX{{\rm B\kern-.05em{\sc i\kern-.025em b}\kern-.08em
    T\kern-.1667em\lower.7ex\hbox{E}\kern-.125emX}}
\usepackage{balance}
\usepackage[colorlinks=true, citecolor=blue]{hyperref}

\begin{document}

\title{\includegraphics[scale=0.02]{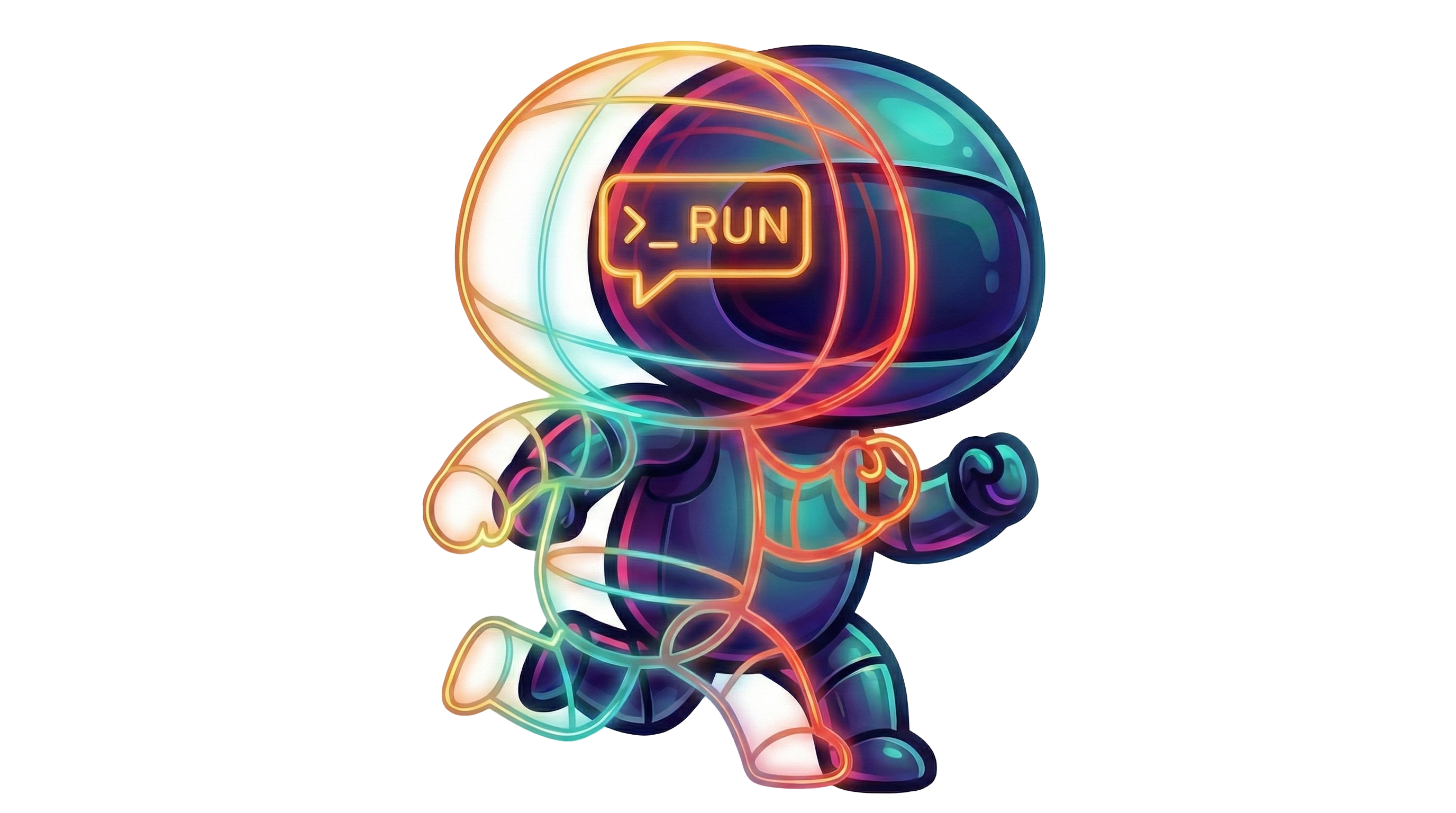}\hspace{-.5cm}Think Before You Move: Latent Motion Reasoning for Text-to-Motion Generation}

\author{Yijie Qian, Juncheng Wang, Yuxiang Feng, Chao Xu, Wang Lu, Yang Liu, \\ Baigui Sun, Yiqiang Chen, \emph{Fellow}, IEEE, Yong Liu, \emph{Member}, IEEE, Shujun Wang, \emph{Member}, IEEE\thanks{Corresponding to: Yong Liu, Shujun Wang; Yijie Qian and Juncheng Wang contribute equally in this project.\\
Yijie Qian, Yuxiang Feng, Yong Liu are with Zhejiang University, Hangzhou, China. Email: yijieqian@zju.edu.cn, fengyx1026@gamil.com, yongliu@iipc.zju.edu.cn.\\
Juncheng Wang, Shujun Wang are with the Hong Kong Polytechnic University, Hong Kong, China. Email: wjc2830@gmail.com, shu-jun.wang@polyu.edu.hk.\\
Chao Xu, and Baigui Sun are with IROOTECH TECHNOLOGY and Wolf 1069 b Lab, Sany Group, Hangzhou, China. Email: chaoxuxc@gmail.com,
sunbaigui85@gmail.com.\\
Yang Liu is with IROOTECH TECHNOLOGY; King’s College London, London, United Kingdom and Wolf 1069 b Lab, Sany Group, Hangzhou, China. Email: yang.15.liu@kcl.ac.uk\\
Wang Lu, and Yiqiang Chen are with Institute of Computing Technology, Chinese Academy of Sciences, Beijing, China. Email: newlw230630@gmail.com, yqchen@ict.ac.cn.
}
}
\markboth{ARXIV PREPRINT}%
{Qian \MakeLowercase{\textit{et al.}}: Think Before You Move: Latent Motion Reasoning for Text-to-Motion Generation}

\IEEEtitleabstractindextext{%
\begin{abstract}
Current state-of-the-art paradigms predominantly treat Text-to-Motion (T2M) generation as a direct translation problem, mapping symbolic language directly to continuous poses. While effective for simple actions, this ``System 1'' approach faces a fundamental theoretical bottleneck we identify as the \textbf{Semantic-Kinematic Impedance Mismatch:} the inherent difficulty of grounding semantically dense, discrete linguistic intent into kinematically dense, high-frequency motion data in a single shot. In this paper, we argue that the solution lies in an architectural shift towards \textbf{Latent System 2 Reasoning}. Drawing inspiration from Hierarchical Motor Control in cognitive science, we propose Latent Motion Reasoning (LMR) that reformulates generation as a two-stage ``Think-then-Act'' decision process.
Central to LMR is a novel \textbf{Dual-Granularity Tokenizer} that disentangles motion into two distinct manifolds: a compressed, semantically rich Reasoning Latent for planning global topology, and a high-frequency Execution Latent for preserving physical fidelity. By forcing the model to autoregressively ``reason'' (plan the coarse trajectory) before it ``moves'' (instantiates the frames), we effectively bridge the ineffability gap between language and physics. We demonstrate LMR's versatility by implementing it for two representative baselines: T2M-GPT (discrete) and MotionStreamer (continuous). Extensive experiments show that LMR yields non-trivial improvements in both semantic alignment and physical plausibility, validating that the optimal substrate for motion planning is not natural language, but a learned, motion-aligned concept space. Codes and demos can be found in \href{https://chenhaoqcdyq.github.io/LMR/}{https://chenhaoqcdyq.github.io/LMR/}.

\end{abstract}

}

\maketitle

\IEEEpeerreviewmaketitle

\section{Introduction}
\input{Sections/1_new_introduction}
\input{Sections/2_new_related_works}
\input{Sections/2_5_new_findings}
\input{Sections/3_new_method}
\input{Sections/4_new_experiments}
\input{Sections/5_conclusion}

{
    \small
    \bibliographystyle{IEEEtran}
    \bibliography{IEEEabrv,new_ref}
}

\clearpage
\appendices
\makeatletter
\renewcommand{\@IEEEprocessthesectionargument}[1]{%
  \@ifmtarg{#1}{%
    \@IEEEappendixsavesection*{\appendixname\nobreakspace\thesectiondis}%
    \addcontentsline{toc}{section}{\appendixname\nobreakspace\thesection}}{%
    \@IEEEappendixsavesection*{\thesectiondis. #1}%
    \addcontentsline{toc}{section}{\appendixname\nobreakspace\thesection: #1}}}
\makeatother
\section*{Appendices}
\input{Sections/appendix}

\end{document}

%% file: Sections/1_new_introduction.tex
The rapid evolution of generative models has transformed Text-to-Motion (T2M) generation~\cite{guo2022generating, MDM, T2M-GPT, MLD, MotionLCM} from a niche subfield into a critical frontier for computer animation and embodied AI. Current state-of-the-art paradigms predominantly treat T2M as a direct translation problem: a sequence-to-sequence mapping where a source modality (natural language) is projected directly onto a target modality (joint rotations or positions)~\cite{PhysDiff, MoFusion}.
\input{NewFigures/Fig_Teaser}
While ``System 1''~\cite{kahneman2011thinking} approaches (Fig.~\ref{fig:teaser_fig}-(a),(b)) achieve success in simple tasks via reflexive feed-forward mapping, they are constrained by the \textbf{Semantic-Kinematic Impedance Mismatch}, which hinders the accurate modeling of physically nuanced dynamics~\cite{HuMoR, SimPoE,pouw2021semantically}.

This impedance mismatch fundamentally stems from the discrepancy between natural language and raw motion data. The former is discrete, symbolic, and semantically dense~\cite{li2025integrating}; it captures high-level intent (``walk dejectedly'') but discards the low-level, high-frequency continuous information required for execution (velocity profiles, center-of-mass shifts, and foot-contact timing).
Conversely, the latter is continuous, real-valued, and kinematically dense yet semantically sparse. Bridging this gap with a single neural network requires inferring precise physical dynamics directly from abstract linguistic descriptors, necessitating the simultaneous resolution of two orthogonal objectives: semantic alignment and physical control~\cite{OmniControl, GMD}. The result is often ``gliding,'' physical implausibility, or generic motion that lacks the nuanced style implied by the text.

We argue that the solution lies not in advanced tokenization representations or generative paradigms, but in a fundamental architectural shift towards \textbf{Latent System 2 Reasoning}~\cite{kahneman2011thinking,booch2021thinking,li2025system}.
In cognitive science and motor neuroscience, biological systems do not map high-level goals directly to muscle activations~\cite{egan2015role,latash2008evolution, czyz2019development}. Instead, biological agents adopt a \textbf{Hierarchical Motor Control} strategy that synthesizes a ``Generalized Motor Program'' (GMP)~\cite{carter1984control, Merel2019, CALM} to serve as an abstract latent representation of movement intent, which is subsequently modulated into specific neuromuscular execution commands. This intermediate stage acts as a cognitive buffer, a ``reasoning space'' where the system resolves ambiguity and plans the trajectory topology before committing to physical execution. Inspired by this, Chain of Thought (CoT)~\cite{wei2022chain} and Latent Reasoning~\cite{hao2024training} have sparked revolutionary advances in Large Language Models (LLMs).

Consequently, we propose that T2M generation requires a similar architectural evolution: the introduction of an intermediate reasoning phase. However, we argue that a direct adaptation relying on \textbf{explicit language reasoning} (Fig.~\ref{fig:teaser_fig}-(c)), such as the textual intermediate steps employed by Motion-R1~\cite{MotionR1}, is inherently ill-suited for the continuous nature of motion tasks~\cite{ChatMotion}.
This limitation stems from the \textbf{inherent ineffability} of physical dynamics. While natural language excels at semantic abstraction, it acts as a lossy compression algorithm for kinematics; describing the precise transition of weight or the subtle damping of a joint during a dance move is verbose and imprecise in English, yet geometrically distinct in a high-dimensional vector space. Therefore, restricting the reasoning process to the linguistic domain forces the model to discard essential high-frequency signal, limiting its planning capabilities to \textbf{language fluency} while failing to ensure \textbf{physical validity}.

In this paper, we posit that the optimal substrate for this planning phase is not natural language, but a latent modality (Fig~\ref{fig:teaser_fig}-{(d)}). By introducing an intermediate latent reasoning stage—$p(\texttt{latent} | \texttt{text})$—we allow the model to engage with a rich, intermediate abstraction that captures the physics and style of the motion before expanding it into raw frames via $p(\texttt{motion} | \texttt{latent})$. This paradigm shifts T2M from a task of translation to one of \textbf{progressive instantiation}, bridging the ineffability gap by allowing the model to think in motion-aligned concepts before it acts.

This shift prompts a critical inquiry: \textbf{what constitutes an effective latent substrate that balances semantic tractability} (i.e., a representation easy for the model to reason over and align with text) \textbf{with physical fidelity} (i.e., motion that is physically plausible as real human movement)\textbf{?} To answer this, our research commences with empirical studies probing the nature of the feature spaces of motion token sequences.
\underline{Firstly}, we analyze the manifolds induced by different pre-training objectives. We observe a distinct orthogonality: latent spaces optimized for semantic alignment efficiently capture global temporal structure (the \emph{what} and \emph{why}), while those optimized for reconstruction capture local geometric precision (the \emph{how}).
This reveals that \emph{the learning objective of tokenizers determines the capacity of semantic or kinematic knowledge} (``knowledge capacity'').
\underline{Secondly}, we re-examine the established trade-off in tokenization. While conventional wisdom suggests that longer token sequences degrade generation due to accumulation error, our empirical observations reveal a more nuanced reality: the bottleneck is not sequence length per se, but the compression rate. Low-compression tokenizers achieve better reconstruction by yielding ``kinematically dense'' tokens, but worse generation because those tokens are ``semantically sparse''; the converse holds for high-compression tokenizers.
Hence, \emph{the compression rate of tokenizer determines how the semantic or kinematic knowledge is distributed over tokens} (``knowledge density'').

Guided by these insights, we instantiate \textbf{Latent Motion Reasoning} (LMR), a framework that reformulates autoregressive T2M generation from a flat sequence modeling task into a \textbf{hierarchical decision process}.
At its core, LMR abandons the monolithic latent assumption in favor of a \textbf{Dual-Granularity Tokenization} strategy.
We explicitly disentangle the representation into two manifolds: a \textbf{Reasoning Latent} and an \textbf{Execution Latent}.
Specifically, the \textbf{Reasoning Latent} is fed with raw motion data (to ensure kinematic capacity) and aligned with semantically rich textual embeddings (to ensure semantic capacity); it is further projected through a tailored compression rate to modulate the balance of density between kinematic and semantic knowledge.
The \textbf{Execution Latent}, in contrast, is optimized purely for reconstruction to preserve high-frequency motion fidelity, capturing the motion's ``dynamics''.
Consequently, we restructure the generation horizon into a causal chain: \texttt{text} $\to$ \texttt{reasoning (plan)} $\to$ \texttt{execution (act)}. By training the generator to predict this structured sequence, LMR first hallucinates the coarse-grained topology of the movement before ``filling in'' the precise physical details, effectively decoupling motion planning from motion instantiation so the model can ``think'' in stable concepts before committing to volatile frames.
This core integrates naturally with both discrete and continuous tokenizations.

To demonstrate the versatility of LMR, we instantiate it on two representative autoregressive T2M baselines: T2M-GPT~\cite{T2M-GPT} (discrete token space, cross-entropy loss) and MotionStreamer~\cite{MotionStreamer} (continuous token space, diffusion loss). Extensive experiments demonstrate substantial improvements over these backbones, specifically achieving 71\% and 64\% reductions in FID on the HumanML3D~\cite{HumanML3D} and KIT-ML~\cite{KIT-ML} datasets for T2M-GPT, respectively, along with a 15\% improvement over MotionStreamer on HumanML3D.

Our contributions are four-fold:
\begin{itemize}
    \item \textbf{Concept}: We identify the Semantic-Kinematic Impedance Mismatch in T2M and propose Latent System 2 Reasoning as a solution, arguing that motion planning is inherently non-linguistic.
    \item \textbf{Analysis}: We reveal two properties in motion tokenization, empirically demonstrating the knowledge capacity and density between semantic alignment (reasoning) and kinematic fidelity (execution) in token sequences.
    \item \textbf{Method}: We propose LMR, a latent reasoning framework utilizing a dual-granularity tokenizer that decouples motion planning from motion execution.
    \item \textbf{Performance}: We demonstrate that LMR significantly boosts the performance of existing state-of-the-art models (T2M-GPT and MotionStreamer), validating the efficacy of the "think before you move" paradigm.
\end{itemize}

%% file: NewFigures/Fig_Teaser.tex
\begin{figure}
    \centering
    \includegraphics[width=\linewidth]{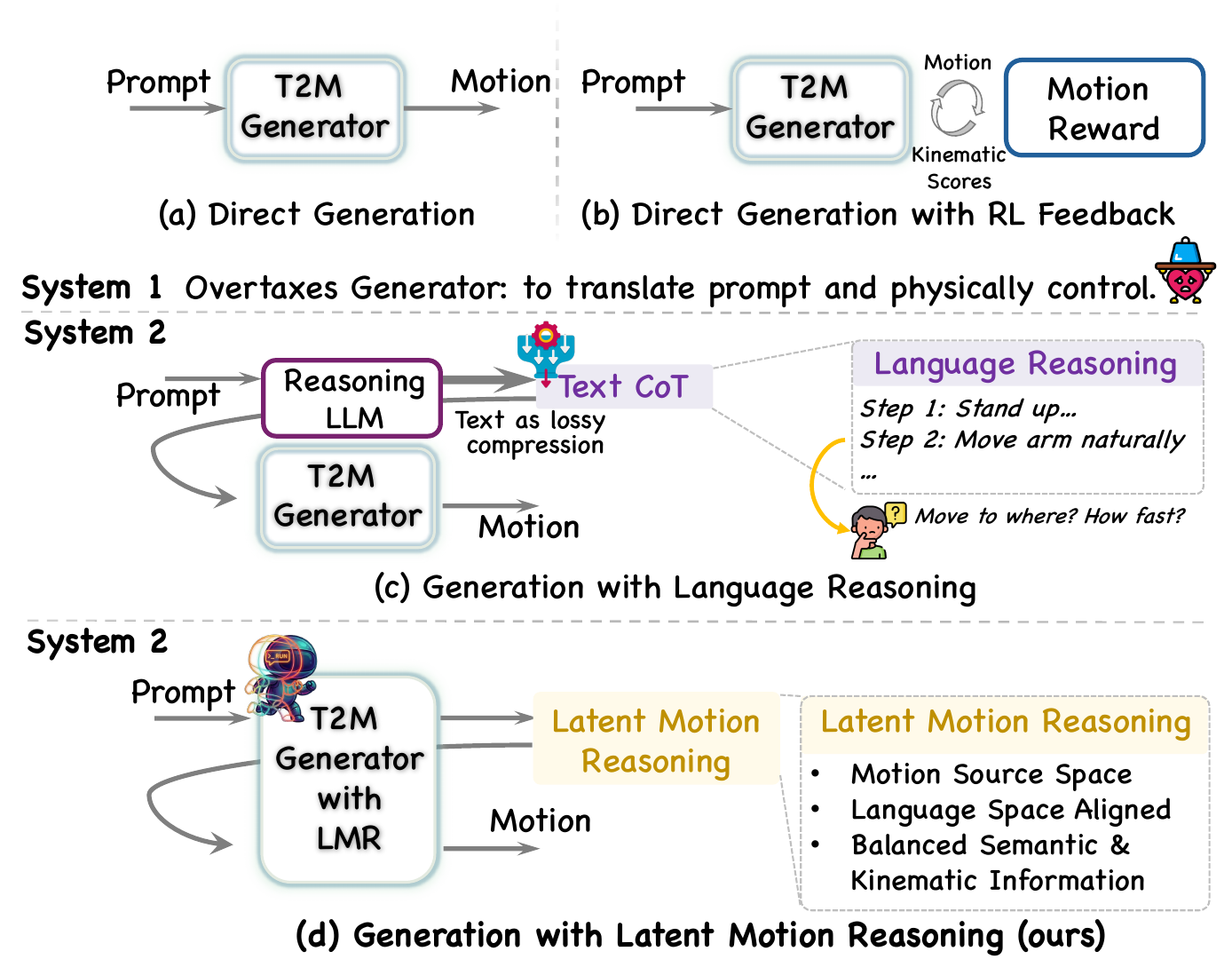}
    \caption{\textbf{Architectural Comparison and the Central Challenge of Text-to-Motion (T2M) Generation}. Existing methods struggle to bridge the gap between abstract language and continuous motion. \textbf{(a) Direct Generation}~\cite{T2M,MotionStreamer} and \textbf{(b) Direct Generation with Feedback}~\cite{MotionRL,yue2025rl} (collectively, System 1) are severely limited by the \textbf{Semantic-Kinematic Impedance}, where the generator must simultaneously plan global trajectory and guarantee local physical fidelity, resulting in a single-shot, over-burdened process and outputs that lack physical grounding. \textbf{(c) The Language CoT Paradigm}~\cite{MotionR1} introduces high-level reasoning (System 2), but the textual chain-of-thought is a Low-Bandwidth Bottleneck (Funnel), losing the high-frequency information necessary for realistic dynamics—a limitation known as the Ineffability of Language. \textbf{(d) Our Latent Motion Reasoning (LMR)} overcomes these limitations by operating entirely within the Dual-Granularity Latent Space. This ``Think-then-Act'' architectural shift disentangles planning from execution, allowing the model to \textbf{autoregressively} reason over a compressed trajectory before instantiating the \textbf{physically plausible} motion, effectively harmonizing semantic intent with kinematic demands.}
    \label{fig:teaser_fig}
\end{figure}

%% file: Sections/2_new_related_works.tex
\section{Related Works}

\subsection{Text-to-Motion Generation}
The field of text-to-motion generation has evolved from early deterministic approaches to probabilistic generative frameworks. Early efforts~\cite{cai2018deep, wang2020learning, petrovich2021action, guo2022generating} employed GANs~\cite{GAN} and VAEs~\cite{VAE}, while others explored joint embedding spaces for better language grounding~\cite{Language2Pose, JL2P}. However, these methods often struggled with the diversity and quality of long-term motion.

\vspace{0.3em}
\noindent\textbf{Diffusion-based Methods.} With the success of diffusion models in image synthesis, methods like MDM~\cite{MDM} and MotionDiffuse~\cite{MotionDiffuse} introduced diffusion processes to raw motion sequences. MotionCLIP~\cite{MotionCLIP} demonstrated the efficacy of aligning motion with CLIP space. MLD~\cite{MLD} improved efficiency by performing diffusion in a learned latent space. Subsequent works have focused on specific challenges: MotionLCM~\cite{MotionLCM} optimizes runtime efficiency; ReMoDiffuse~\cite{ReMoDiffuse} enhances diversity; DiffCollage~\cite{DiffCollage} enables long-term generation via parallel diffusion; HumanTOMATO~\cite{HumanTOMATO} focuses on whole-body motion alignment; Fg-T2M~\cite{Fg-T2M} targets fine-grained semantic alignment; and MARDM~\cite{MARDM} revisits diffusion by integrating it with masked autoregression to reduce representation redundancy.

\vspace{0.3em}
\noindent\textbf{Discrete Token-based Methods.} Another prominent paradigm involves discretizing motion into codebook indices, typically following two architectural branches:
\begin{itemize}
    \item \textbf{Masked Modeling (BERT-style):} Methods like MoMask~\cite{MoMask} and MMM~\cite{MMM} employ bidirectional transformers to reconstruct masked tokens. While effective, they typically require ground-truth motion length as a prior. To mitigate this, the concurrent BAMM~\cite{BAMM} proposes a hybrid attention masking strategy to unify bidirectional context with autoregressive length prediction.
    \item \textbf{Autoregressive Modeling (GPT-style):} These methods predict tokens sequentially. T2M-GPT~\cite{T2M-GPT} established the baseline using VQ-VAE and standard GPT. Recent advancements focus on scalability and streaming. AttT2M~\cite{AttT2M} enhances semantic alignment via multi-perspective attention. MotionStreamer~\cite{MotionStreamer} enables streaming generation through causal constraints. MOGO~\cite{Mogo} introduces a hierarchical causal transformer for infinite-length generation. Similarly, MoSa~\cite{mosa} proposes scalable autoregressive modeling with multi-scale token supervision to improve efficiency.
\end{itemize}

Despite these advancements, a fundamental theoretical bottleneck persists: the \textbf{Semantic-Kinematic Impedance Mismatch}. Current state-of-the-art paradigms predominantly treat T2M as a ``System 1'' direct translation task, attempting to map discrete, symbolic linguistic intent onto kinematically dense, high-frequency pose data in a single pass. While effective for reflexive actions, this approach struggles to resolve the simultaneous objectives of high-level semantic alignment and low-level physical control. Unlike prior works that aggressively downsample motion to mitigate modeling complexity—often at the cost of physical fidelity —we argue that the solution lies in an architectural decoupling. Our LMR framework reformulates generation as a hierarchical "Think-then-Act" process, introducing a latent reasoning buffer that allows the model to bridge the ineffability gap through structured planning before committing to kinematic execution.
\subsection{Motion Tokenizer}
The quality of motion generation is fundamentally bounded by the tokenizer's ability to compress and reconstruct dynamics.
\textbf{Standard VQ-VAE approaches}~\cite{TM2T, T2M-GPT} map motion snippets to single codebook indices but often suffer from "codebook collapse" or loss of fine details.
\textbf{Residual Quantization (RVQ)} has been adopted by MoMask~\cite{MoMask} and MOGO~\cite{Mogo} to reduce quantization error by using multiple codebooks sequentially.
\textbf{Structure-aware approaches} like ParCo~\cite{ParCo} decompose motion into body parts to learn distinct codebooks.
Recently, \textbf{Hierarchical designs} have emerged: MoSa~\cite{mosa} utilizes a multi-scale token preservation strategy to balance global structure and local detail.
Besides, due to the lossy compression of quantization based tokenizers, motionstreamer~\cite{MotionStreamer} conducts next-token-prediction in continuous space with diffusively predicting a latent by KL-VAE~\cite{VAE}.

In summary, the efficacy of motion generation is intrinsically bounded by the tokenizer’s ability to preserve the delicate balance between semantic tractability and kinematic precision. We identify that conventional monolithic tokenizers often fall into two extremes: ``Semantic Sparsity'', where high-frequency details dilute the planning signal, or ``Kinematic Loss'', where excessive compression discards the nuanced dynamics required for realism. Our approach moves beyond simple temporal downsampling. By implementing a Dual-Granularity Tokenizer, we explicitly disentangle the latent space into two manifolds: a Reasoning Latent optimized for global topology and an Execution Latent reserved for physical fidelity. This design ensures a robust, motion-aligned foundation for the subsequent reasoning module, effectively addressing the representation gap inherent in existing methods.

\subsection{Evolution of Chain-of-Thought: From Linguistic to Latent Reasoning}
The paradigm of Chain-of-Thought (CoT) prompting has revolutionized Large Language Models (LLMs) by decomposing complex problems into intermediate, interpretable reasoning steps~\cite{CoT}. This cognitive ``System 2'' shift has rapidly transcended natural language, expanding into multimodal domains to address challenges in spatial control and structural synthesis. In the visual domain, frameworks such as Visual CoT~\cite{visualcot, harvey2023visual} and VCTP~\cite{chen2024visual, rose2023visual} mimic human deduction through iterative inspection. Similarly, layout-based reasoning~\cite{feng2023layoutgpt, song2023llm, srivastava2025lay, li2025llms} and the ``Thinking with Images'' paradigm~\cite{chern2025thinking, su2025thinking} utilize intermediate symbolic layouts or low-fidelity drafts to facilitate pixel synthesis. Beyond vision, CoT has been employed in music~\cite{copet2023simple, lam2025analyzable} and audio generation~\cite{wang2025language, liuthinksound, wang2025synchronized} to impose high-level structural logic upon low-level waveforms.

However, a fundamental discrepancy remains in how reasoning is grounded when transitioning from static semantics to dynamic physics. In the specific context of motion generation, recent attempts like Motion-R1~\cite{MotionR1} and CoT-Pose~\cite{cotpose} have introduced CoT by decomposing prompts into explicit text-based sub-steps (e.g., planning "stance" before "movement"). While these approaches enhance interpretability, we argue they are ultimately constrained by the \textit{Ineffability of Language}. Natural language, being discrete and symbolic, acts as a lossy, low-bandwidth bottleneck that discards the high-frequency kinematic details---such as velocity profiles and momentum shifts---essential for realistic execution~\cite{chern2025thinking,su2025thinking}.

Our proposed \textbf{Latent Motion Reasoning (LMR)} represents a pivotal departure from these linguistic-centric paradigms. Instead of forcing physical dynamics into the narrow confines of text, LMR implements CoT reasoning within a motion-aligned latent space. By operating in this dual-granularity manifold, the model can ``think'' using high-dimensional concepts that are semantically dense yet kinematically grounded. This shift effectively bridges the Semantic-Kinematic Impedance Mismatch: rather than a mere translation of symbols, T2M generation becomes a process of progressive instantiation, where the model resolves global trajectory topology in the reasoning latent before committing to volatile physical frames.

%% file: Sections/2_5_new_findings.tex
\section{Preliminaries of Text-to-Motion Generation}
\label{sec:preliminaries}

In this section, we formulate the text-to-motion (T2M) task and review the standard autoregressive frameworks that serve as the backbone for our proposed method.

\subsection{Problem Formulation}
Let $\mathbf{x} = [x_1, x_2, \dots, x_{T_m}]$ represent a sequence of motion frames, where each $x_t \in \mathbb{R}^d$ describes the pose (e.g., joint rotations and root trajectory) at time step $t$. Given a natural language description $c$, the goal of T2M generation is to model the conditional distribution $p(\mathbf{x}|c)$.
To leverage the semantic knowledge of large-scale language models, the text prompt $c$ is typically encoded into a static embedding using a pre-trained encoder such as CLIP~\cite{CLIP}.

\input{NewFigures/Fig_Findings_1}
\subsection{Motion Tokenization}
Due to the high dimensionality and redundancy of raw motion data, state-of-the-art approaches typically compress $\mathbf{x}$ into a lower-dimensional latent sequence before generation. This is achieved via an Encoder-Decoder architecture ($\mathcal{E}, \mathcal{D}$).

\subsubsection{Discrete Quantization (VQ-VAE)}
In discrete frameworks (e.g., T2M-GPT), the motion is mapped to a sequence of codebook indices. The encoder $\mathcal{E}$ maps $\mathbf{x}$ to latent features $f \in \mathbb{R}^{T' \times d'}$, which are then discretized using a learnable codebook $\mathcal{Z} = \{z_k\}_{k=1}^K$. The quantization operation replaces each feature vector with its nearest codebook neighbor:
\begin{equation}
    q = \arg \min_k || f - z_k ||_2
\end{equation}

\subsubsection{Continuous Representation (VAE)}
In continuous frameworks (e.g., MotionStreamer), vector quantization is omitted to preserve arithmetic properties of the latent space. Instead, a Variational Autoencoder (VAE) is often used, where the encoder predicts parameters of a distribution (e.g., $\mu, \sigma$) from which the latent tokens are sampled. This effectively projects the motion onto a continuous manifold bounded by a regularization term, such as KL-divergence.

\subsection{Standard Autoregressive Modeling}
Standard autoregressive (AR) models decompose the generation of the motion sequence $S$ into a product of conditional probabilities:
\begin{equation}
    p(S|c) = \prod_{t=1}^{T'} p(S_t | S_{<t}, c)
\end{equation}
A Transformer backbone $\mathcal{H}_\theta$ processes the text condition $c$ and the history of generated tokens $S_{<t}$ to produce a hidden state $h_t = \mathcal{H}_\theta(S_{<t}, c)$ at each time step $t$. The modeling of the distribution $p(S_t | \cdot)$ depends on the nature of the latent space:

\subsubsection{Discrete Autoregression (Classification)}
In discrete approaches (e.g., T2M-GPT), $S_t$ is a codebook index. The hidden state $h_t$ is projected via a linear head to a categorical distribution, and the model is trained via Cross-Entropy loss to predict the next token index.

\subsubsection{Continuous Autoregression (AR-Diffusion)}
In continuous approaches (e.g., MotionStreamer), $S_t \in \mathbb{R}^{d'}$ is a continuous vector. Rather than regressing $S_t$ directly (which often leads to mean-collapse), the distribution $p(S_t | S_{<t}, c)$ is modeled by a conditional diffusion process.
At each temporal step $t$, the AR hidden state $h_t$ serves as the condition for a denoising network $\epsilon_\phi$. The token $S_t$ is generated by iteratively denoising a Gaussian noise sample $S_t^{(L)} \sim \mathcal{N}(0, I)$ over $L$ diffusion timesteps:
\begin{equation}
    S_t^{(l-1)} = \text{Denoise}(S_t^{(l)}, l, h_t)
\end{equation}
During training, this is optimized via a Denoising Score Matching objective applied at every sequence position $t$:
\begin{equation}
    \mathcal{L}_{diff} = \mathbb{E}_{t, l, \epsilon} [ || \epsilon - \epsilon_\phi(S_t^{(l)}, l, h_t) ||_2^2 ]
\end{equation}
where $S_t^{(l)}$ is the noisy version of the ground truth token at diffusion step $l$. This paradigm requires running the full diffusion reverse process to instantiate the token at step $t$ before the AR model can proceed to step $t+1$.
Crucially, these standard approaches map semantic intent $c$ directly to kinematic execution $S$ in a single pass, creating the \textit{Semantic-Kinematic Impedance Mismatch} discussed in introduction section.

\section{On the Knowledge Capacity and Density in Token Spaces}
\label{sec:semantic_density}
To address the \textit{Semantic-Kinematic Impedance Mismatch}, this paper advocates for a ``System 2'' architectural shift, decomposing generation into high-level planning and low-level execution. However, implementing this paradigm necessitates a specific representational substrate: a latent space that is sufficiently abstract to align with linguistic intent, yet sufficiently grounded to control physical kinematics.

We posit that the optimality of such a substrate is defined by a delicate \textbf{Knowledge Balance} between semantic abstraction and kinematic fidelity. To systematically analyze this balance, we formulate two governing properties of motion tokenization:
\begin{itemize}
\item \textbf{Knowledge Capacity (The Manifold):} This refers to the intrinsic scope of information the latent manifold is capable of representing. We hypothesize that the \textit{type} of knowledge a tokenizer prioritizes—whether global semantics or local geometry—is fundamentally determined by its \textbf{pre-training objectives}.
\item \textbf{Knowledge Density (The Sequence):} This refers to the concentration of information allocated to each individual token in the autoregressive sequence. We hypothesize that the \textit{richness} of the reasoning signal versus the sparsity of the execution detail is directly dictated by the tokenizer's \textbf{temporal compression rate}.
\end{itemize}
To validate this framework, we conduct two pilot studies to reveal how the training objectives (Sec.~\ref{sec:learning_objective}) and compression rates (Sec.~\ref{sec:tokenization_rate}) serve as the distinct control levers for Capacity and Density, respectively.

\subsection{Pre-Training Objectives: The Manifold Orthogonality}
\label{sec:learning_objective}
To validate how pre-training objectives govern the manifold, we analyze the latent geometries induced by the two dominant pre-training objectives in T2M: \textit{Reconstruction} ($\mathcal{L}_{rec}$) \cite{VAE} and \textit{Semantic Alignment} ($\mathcal{L}_{align}$)~\cite{TMR}.

We train two distinct motion tokenizer variants on the HumanML3D dataset. Model A (VQ-VAE) is trained with reconstruction objective $\mathcal{L}_{rec}$ (MSE), while Model B (TMR Encoder~\cite{TMR}) uses a contrastive objective $\mathcal{L}_{align}$, aligning motion features with CLIP text embeddings. We probe these latent spaces via two distributions.
Given the same tokenized motion sequences, we visualize the cluster distributions according to two distinct label sets, one label set classifies sequences according to the semantic meaning (such as ``interactions'', ``object handling''), while another label distinguishes sequences as the kinematic actions (such as those with similar joint angles).

As visualized in Fig.~\ref{fig:tsne_manifold}, t-SNE projections reveal a fundamental orthogonality in how these objectives structure the latent space:
\begin{itemize}
    \item \textbf{Semantic Clustering ($\mathcal{L}_{align}$):} The alignment-optimized space organizes data by intent (as shown by the second column of Fig.~\ref{fig:tsne_manifold}). Distinct sequences that share a semantic label (e.g., ``boxing'' punch vs. ``boxing'' dodge) are drawn together, while sequences with the same action labels seem to be randomly distributed.
    \item \textbf{Kinematic Clustering ($\mathcal{L}_{rec}$):}
    The reconstruction-optimized space organizes data by geometric similarity (as shown by the first column of Fig.~\ref{fig:tsne_manifold}).
    Motions with similar kinematic dynamics are clustered tightly. However, semantic concepts (e.g., ``interactions'' vs. ``walking'') are entangled.
\end{itemize}

\begin{tcolorbox}[
    arc=5pt,          
    colback=blue!5!white, 
    colframe=blue!75!black, 
    boxrule=1pt,      
    sharp corners=northwest, 
]
\textbf{Takeaway 1: The Necessity of Manifold Disentanglement}.
These findings confirm our hypothesis regarding \textbf{Knowledge Capacity}: we observe a fundamental ``Manifold Orthogonality'' where latent spaces optimized for kinematic reconstruction $(\mathcal{L}_{rec})$ and semantic alignment $(\mathcal{L}_{align})$ evolve into mutually exclusive topologies. This indicates that a monolithic latent space suffers from an inherent capacity bottleneck—it cannot simultaneously accommodate the high-frequency geometric precision required for execution and the abstract semantic clustering required for reasoning. Consequently, the optimal substrate is not a compromise, but a decoupling. This motivates the \textbf{Manifold Disentanglement} in our Dual-Granularity Tokenizer (Sec.~\ref{sec:method}), which assigns distinct subspaces to maximize the capacity for semantic planning (``the what'') and kinematic instantiation (``the how'') independently, and then aggregates the knowledge capacity in a shared encoder.
\end{tcolorbox}

\subsection{Tokenization Temporal Compression Rate}
\label{sec:tokenization_rate}
\input{NewFigures/Fig_Findings_2}
Complementing our analysis of pre-training objectives, we now examine the \textit{temporal resolution} of the latent space. While the learning objective determines \textit{what} features are preserved, the tokenization rate dictates their \textit{distribution}. We identify a fundamental tension: high-frequency tokenizers (with less compression rate) minimize reconstruction error but dilute semantic signals, creating a ``contextual sprawl'' that hinders autoregressive (AR) modeling. This paradox compels us to investigate the trade-off between physical fidelity and sequence modeling tractability.

\subsubsection{Sequence Length Trade-off Analysis}
Given a motion sequence with a length of $T_m$, if we increase the token sequence $T$, it naturally improves reconstruction but exacerbates long-term dependencies. To probe this, we evaluate VQ-VAE configurations across varying downsampling rates ($T \in T_m \times \{1, 1/2, 1/4, 1/8\}$). As shown in Fig.~\ref{fig:manifold_tradeoff}, we observe a non-monotonic trend: while reconstruction (MPJPE) improves linearly with length, generation quality (gFID) plateaus at $T = T_m/4$. This suggests that simply increasing resolution yields diminishing returns for generation due to modeling complexity.

\subsubsection{The Semantic Density Hypothesis}
To isolate the impact of temporal resolution, we further utilize two motion sequences, one with a length of $T_m$, another with $4\times T_m$ (by up-sampling the same $T_m$ motion). For $T_m$ motion, we utilize a high-frequency tokenizer (no temporal compression).
As for $4\times T_m$ motion, we utilize a relatively low-frequency tokenizer (four times temporal compression).
With such, we can derive two motion sequences with the same length.

Comparing the generation performance over these two token sequences, we observe that \emph{the longer motion sequence achieves better generation FID, even though it is harder to reconstruct}.
This confirms that the bottleneck is not sequence length but \textbf{Knowledge Density}—the ratio of information content per token.
High-frequency sequences dilute semantic intent across many frames while preserving more kinematic detail; the converse holds for low-frequency sequences.
This is further validated by our ablation studies (middle \& right columns of Fig.~\ref{fig:manifold_tradeoff}): randomly dropping tokens in longer sequences causes significantly lower retrieval degradation than in shorter ones, proving that semantic information is highly redundant and distributed in high-frequency regimes.

\begin{tcolorbox}[
    arc=5pt,          
    colback=blue!5!white, 
    colframe=blue!75!black, 
    boxrule=1pt,      
    sharp corners=northwest, 
]
\textbf{Takeaway 2: Bridging the Granularity Gap via Latent Buffering.}
These findings substantiate the critical role of \textbf{Knowledge Density} in autoregressive modeling. We observe that high-frequency tokenization induces a ``contextual sprawl'' that dilutes semantic intent, resulting in prediction ambiguity within flat autoregressive frameworks. To construct a \textbf{Knowledge Balanced} latent reasoning sequence, it is imperative to decouple the temporal resolution of planning from execution. This insight directly motivates the temporal abstraction strategy in our LMR, which enforces a distinct, compressed sequence length for the reasoning phase to maximize knowledge density.
\end{tcolorbox}

%% file: NewFigures/Fig_Findings_1.tex
\begin{figure*}
    \centering
    \includegraphics[width=.7\linewidth]{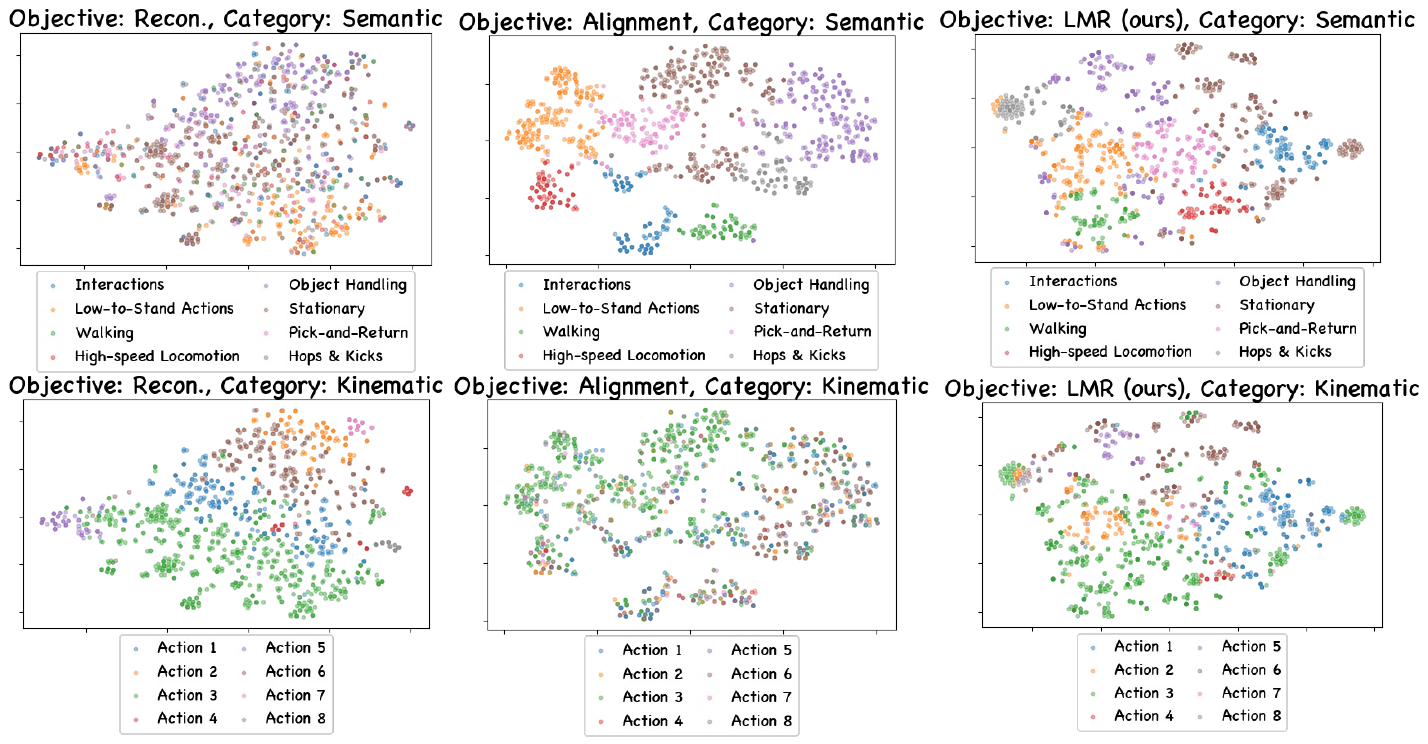}
    \caption{\textbf{Visualization of the Manifold Orthogonality between Semantic Alignment and Kinematic Fidelity}. We employ t-SNE to project motion representations learned under three distinct objectives: \textbf{(Left) Reconstruction-only ($\mathcal{L}_{rec}$), (Middle) Alignment-only ($\mathcal{L}_{align}$), and (Right) Our LMR Framework}. The Top Row colors samples by high-level semantic categories (e.g., Interactions, Object Handling), while the \textbf{Bottom Row} colors them by specific kinematic sequences, which is ineffability, thus denoted as Action X.
    \textbf{Observation}: A fundamental trade-off is observed in single-objective baselines. The Reconstruction objective yields tight kinematic clusters (Left-Bottom) but results in \textbf{semantically entangled manifolds} (Left-Top). Conversely, the Alignment objective creates clear semantic boundaries (Middle-Top) but causes \textbf{kinematic blurring} (Middle-Bottom), losing fine-grained physical distinctiveness. \textbf{(Right) Our LMR}, via its Dual-Granularity Tokenizer, successfully reconciles this conflict, maintaining clear separability in both semantic intent and kinematic execution spaces.}
    \label{fig:tsne_manifold}
\end{figure*}

%% file: NewFigures/Fig_Findings_2.tex
\begin{figure*}[t]
    \centering
    \includegraphics[width=0.8\linewidth]{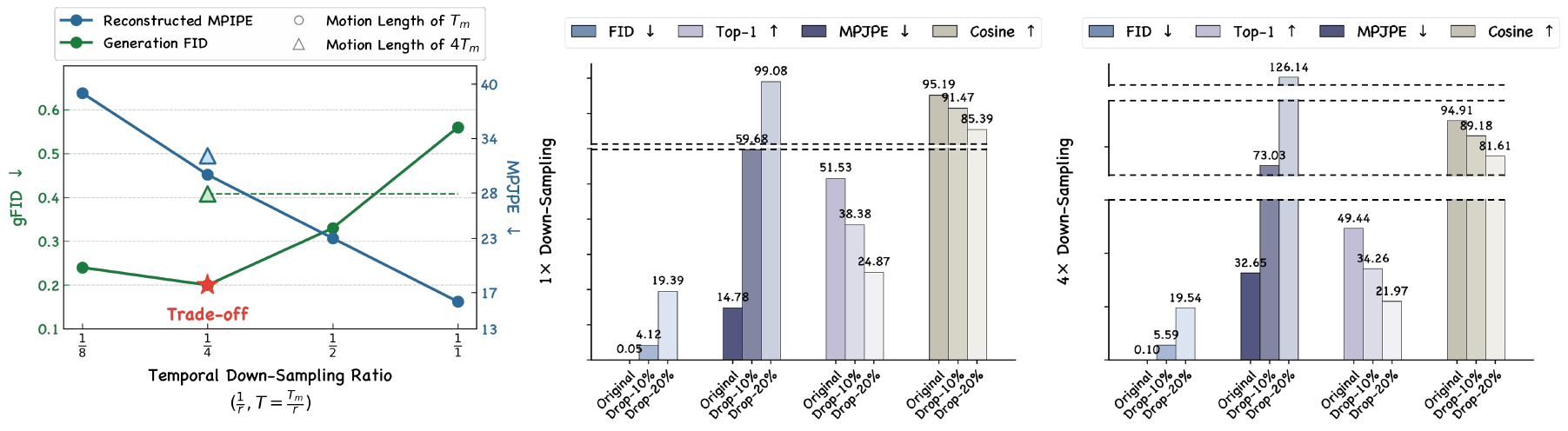}
    \caption{\textbf{Analysis of Semantic Density and Temporal Resolution.
(Left) The Reconstruction-Generation Trade-off}: We plot reconstruction error (MPJPE, Blue) and generation quality (FID, Green) across varying down-sampling ratios. A clear trade-off is observed: while high-frequency tokens ($1/1$) yield the best reconstruction, they degrade generation quality. The optimal balance is found at a $1/4$ ratio (Red Star). Crucially, the ``Same Token Length'' comparison (Triangles vs. Circles with a temporal down-sampling ratio of $\frac{1}{1}$) reveals that a long motion sequence compressed by $4\times$ (Hollow Triangle) generates significantly better motion than a short, uncompressed sequence of the same token count ( $\frac{1}{1}$ Circles). This confirms that \textbf{Semantic Density}, rather than sequence length, is the governing factor for generation quality.
\textbf{(Middle \& Right) Information Sparsity Probe}: We evaluate semantic robustness by testing the retrieval performance using token sequences with randomly dropped tokens. The high-frequency representation (Middle, $1\times$) exhibits higher redundancy (slower decay in Cosine similarity and Top-1 Accuracy) compared to the compressed representation (Right, $4\times$). This indicates that semantic information in raw motion is \textbf{highly sparse and diluted}, creating a ``contextual sprawl'' that hinders effective autoregressive modeling.}
    \label{fig:manifold_tradeoff}
\end{figure*}

%% file: Sections/3_new_method.tex
\section{Latent Motion Reasoning}
\label{sec:method}
Guided by the analysis in Sec.~\ref{sec:semantic_density}, we introduce \textbf{Latent Motion Reasoning (LMR)}. LMR reformulates Text-to-Motion generation as a \textbf{Hierarchical Decision Process}. Instead of mapping text directly to motion tokens, we factorize the generation probability into a two-stage causal chain:
\begin{equation}
p(\mathbf{x} \mid c) \approx
\underbrace{p(S_{{exec}} \mid S_{{res}}, c)}_{\text{Execution (System 1)}}
\cdot
\underbrace{p(S_{{res}} \mid c)}_{\text{Reasoning (System 2)}},
\label{eq:factorization}
\end{equation}
where $S_{res}$ is a highly compressed, semantically dense ``Reasoning Latent'' sequence, and $S_{exec}$ is a high-frequency ``Execution Latent'' sequence.

\subsection{Dual-Granularity Tokenizer}
To materialize the proposed hierarchical reasoning, we design a tokenizer that projects raw motion $\mathbf{x}$ into two disentangled manifolds: a \textbf{Reasoning Manifold} (low-frequency, semantic-dense) and an \textbf{Execution Manifold} (high-frequency, kinematic-dense).
The architecture begins with a shared encoder backbone $\mathcal{E}$ that extracts a base feature sequence $\mathbf{f} = \mathcal{E}(\mathbf{x}) \in \mathbb{R}^{T \times {d'}}$ without temporal compression. From this shared representation, the pipeline bifurcates into two modality-specific implementations.

\subsubsection{Scenario A: Discrete Dual-Codebook Quantization}
In the discrete setting (e.g., for T2M-GPT~\cite{T2M-GPT} baselines), we require the latent space to be categorical. To support the distinct roles of planning and execution, we introduce a \textbf{Dual-Codebook} mechanism comprising a Reasoning Codebook $\mathcal{Z}_{res}$ and an Execution Codebook $\mathcal{Z}_{exec}$.

The Execution Branch (Kinematic VQ):
This branch aims to preserve maximal physical detail. We utilize the full-resolution features $\mathbf{f}$ and quantize them using a large, expressive codebook $\mathcal{Z}_{exec} \in \mathbb{R}^{N_{exec} \times d'}$.
\begin{equation}
    \mathbf{q}_{exec} = \arg \min_{k} \| \mathbf{f} - \mathcal{Z}_{exec}^{(k)} \|_2, \quad \hat{\mathbf{f}}_{exec} = \mathcal{Z}_{exec}(\mathbf{q}_{exec}).
\end{equation}
This yields a sequence $\mathbf{q}_{exec}$ of length $T$, providing the fine-grained tokens required for precise motion reconstruction.

The Reasoning Branch (Semantic VQ):
This branch constructs the ``motor plan.'' We first apply a projection head $\Phi_{res}$ on $\mathbf{f}$, obtaining $\mathbf{f}_{res}$.
Then, we employ a separate, compact codebook $\mathcal{Z}_{res} \in \mathbb{R}^{N_{res} \times d'}$ to discretize these features:
\begin{equation}
     \mathbf{q}_{res} = \arg \min_{k} \| \mathbf{f}_{res} - \mathcal{Z}_{res}^{(k)} \|_2,\quad \hat{\mathbf{f}}_{res} = \mathcal{Z}_{res}(\mathbf{q}_{res}).
\end{equation}
The resulting sequence $\mathbf{q}_{res}$ has length $T/4$. Unlike $\mathcal{Z}_{exec}$ which learns local pose geometry, $\mathcal{Z}_{res}$ is trained to act as a dictionary of ``motion concepts'' or actemes.

\subsubsection{Scenario B: Continuous Dual-Projection}
In the continuous setting (e.g., for MotionStreamer~\cite{MotionStreamer} baselines), quantization is unnecessary. Instead, the focus is on shaping the continuous latent density to be conducive to diffusion processes.

The Execution Branch:
We retain the raw continuous output of the encoder, serving as the ground truth for the diffusion model's reconstruction target:
\begin{equation}
    \mathbf{u}_{exec} = \mathbf{f} \in \mathbb{R}^{T \times d'}.
\end{equation}
We apply a slight KL-regularization penalty $KL(\mathbf{u}_{exec} || \mathcal{N}(0, I))$ to ensure the manifold is bounded, facilitating stable diffusion training.

The Reasoning Branch:
Instead of quantization, we apply an auto-encoder. Similar to the discrete case, we downsample the sequence to $T/4$ by the $\Phi_{res}$ projection head, obtaining continuous latent tokens $\mathbf{u}_{res}$, which serve as the clean condition for the diffusion process.

\subsubsection{Training Objectives for Semantic Density}
Regardless of the modality (Discrete or Continuous), the \textbf{Reasoning Branch} must be forced to capture semantics rather than just compressed kinematics. We impose two auxiliary losses specifically on the reasoning features ($\hat{\mathbf{f}}_{res}$ or $\mathbf{u}_{res}$):

Semantic Alignment (BERT Loss):
To inject language awareness, we maximize the cosine similarity between the pooled reasoning features $\hat{\mathbf{f}}_{res}/\mathbf{u}_{res}$ and the BERT~\cite{BERT} embedding of the text prompt $\mathbf{w}$:
\begin{equation}
    \mathcal{L}_{align} = 1 - \cos(\text{AvgPool}(\mathbf{u}_{res}), \mathbf{w}),
\end{equation}
where $\mathbf{u}_{res}$ illustrates the continuous case; the discrete case uses $\hat{\mathbf{f}}_{res}$ analogously, here and below.

Masked Text Prediction (MTP):
To ensure fine-grained semantic density, we train a lightweight decoder to reconstruct masked words in the text prompt solely from the reasoning tokens:
\begin{equation}
    \mathcal{L}_{mtp} = \text{CrossEntropy}(\text{Decoder}(\mathbf{u}_{res}), \mathbf{w}_{masked}),
\end{equation}
where $\mathbf{w}_{masked}$ denotes the masked language sequence.

The total tokenizer loss is a weighted sum of the reconstruction loss (on the execution branch), the VQ/KL regularization terms, and these semantic objectives.
\begin{equation}
    \mathcal{L}=\mathcal{L}_{rec} + \lambda_{align} * \mathcal{L}_{align} + \lambda_{mtp} * \mathcal{L}_{mtp} + \mathcal{L}_{VQ/KL}.
\end{equation}
This ensures that $\mathbf{u}_{res}$ (or $\hat{\mathbf{f}}_{res}$) emerges as a valid substrate for System 2 reasoning. Intuitively, the two semantic objectives are complementary: $\mathcal{L}_{align}$ enforces \emph{global} text--motion correspondence on the pooled reasoning features, whereas $\mathcal{L}_{mtp}$ grounds \emph{fine-grained} semantics (verbs, body parts, directional words) into individual reasoning tokens.

\input{NewFigures/Fig_Framework}
\subsection{Generative Process: Unified Autoregressive Modeling}
A key advantage of LMR is its architectural flexibility: it integrates seamlessly into existing autoregressive backbones by adopting the native representation space of the host model. Whether the backbone operates on discrete codebook indices (e.g., T2M-GPT) or continuous latent vectors (e.g., MotionStreamer), LMR unifies planning and execution into a \textbf{single heterogeneous sequence}.

We define the generation target as the concatenated sequence $S=[S_{res}; S_{exec}]$ from Eq.~\eqref{eq:factorization}, instantiated in the host backbone's native space as $[\mathbf{q}_{res}; \mathbf{q}_{exec}]$ (discrete) or $[\mathbf{u}_{res}; \mathbf{u}_{exec}]$ (continuous) from the tokenizer. A single Transformer backbone $\mathcal{H}_\theta$ models the joint probability autoregressively, generating the reasoning prefix $S_{res}$ before the execution suffix $S_{exec}$.
We detail the realization of this unified process in two distinct modalities:

\subsubsection{Scenario A: Discrete Token Space (e.g., T2M-GPT)}
In this setting, both the reasoning and execution manifolds are discretized via Vector Quantization~\cite{VQVAE}. The reasoning tokens $\mathbf{q}_{res}$ correspond to indices from the LMR Codebook $\mathcal{Z}_{res}$, while $\mathbf{q}_{exec}$ corresponds to indices from the Execution Codebook $\mathcal{Z}_{exec}$.

The Transformer $\mathcal{H}_\theta$ operates as a standard causal language model. At each step $t$, it predicts the categorical distribution of the next token. We employ a \textbf{Switching Prediction Head} strategy:
\begin{equation}
    p(S_t | S_{<t}, c) =
    \begin{cases}
      \sigma(\mathbf{W}_{res}^T \cdot h_t) & \text{if } t \leq T/4 \text{ (Reasoning Phase)} \\
      \sigma(\mathbf{W}_{exec}^T \cdot h_t) & \text{if } t > T/4 \text{ (Execution Phase)}
    \end{cases}
\end{equation}
where $h_t = \mathcal{H}_\theta(S_{<t}, c)$ is the hidden state, $\sigma$ is the softmax.
Crucially, during the execution phase ($t > T/4$), the self-attention mechanism has full visibility of the completed reasoning sequence $\mathbf{q}_{res}$. This allows the model to attend back to the ``motor plan'' to guide the synthesis of precise motion details, effectively solving the long-horizon forgetting problem via a compact semantic prefix.

\subsubsection{Scenario B: Continuous Feature Space (e.g., MotionStreamer)}
In this setting, we bypass vector quantization during generation. Both reasoning and execution tokens remain in the continuous domain.
Here, $\mathbf{u}_{res}$ consists of the continuous, semantic-aligned features, while $\mathbf{u}_{exec}$ consists of the full-resolution motion features.

We adopt the \textbf{Autoregressive Diffusion} paradigm. The Transformer $\mathcal{H}_\theta$ functions as a conditional denoiser. For each step $t$ in the concatenated sequence, the model synthesizes the continuous vector $S_t$ via a diffusion reverse process.
The training objective is a unified Denoising Score Matching loss:
\begin{equation}
    \mathcal{L}_{diff} = \mathbb{E}_{t, l, \epsilon} \left[ \| \epsilon - \mathcal{H}_\theta(S_t^{(l)}, l, S_{<t}, c) \|_2^2 \right],
\end{equation}
where $l$ is the diffusion timestep and $S_t^{(l)}$ is the noisy version of the current token.

\textbf{Unified Reasoning-Execution Transition:}
Even in this continuous space, the structural advantage remains identical. The model first ``denoises'' the trajectory of the abstract reasoning vectors. Once the planning steps are complete, it transitions to denoising the execution vectors. Because the backbone is autoregressive, the generation of the high-frequency execution features is conditioned on the noise-free, stable trajectory of the reasoning features generated in the first $T/4$ steps.

\subsubsection{Summary of Inference Flow}
In both scenarios, inference proceeds as a single ``Think-then-Act'' stream:
1.  \textbf{Phase I (Reasoning):} The model autoregressively generates the first $T/4$ tokens. These tokens, being semantically dense and temporally compressed, rapidly establish the global topology of the motion.
2.  \textbf{Phase II (Execution):} The model continues generating from $t = T/4 + 1$ to $T/4 + T$. The high-frequency kinematic dynamics are ``filled in'' by attending to the semantic constraints established in Phase I.

This formulation demonstrates that LMR is not tied to a specific generative loss (Cross-Entropy vs. Diffusion) but is rather a universal strategy for \textbf{Hierarchical Temporal Modeling}. The complete tokenizer-training, generator-training, and inference procedures are summarized as pseudo-code in Appendix~\ref{app:pseudocode}.

%% file: NewFigures/Fig_Framework.tex
\begin{figure*}[t]
    \centering
    \includegraphics[width=0.9\linewidth]{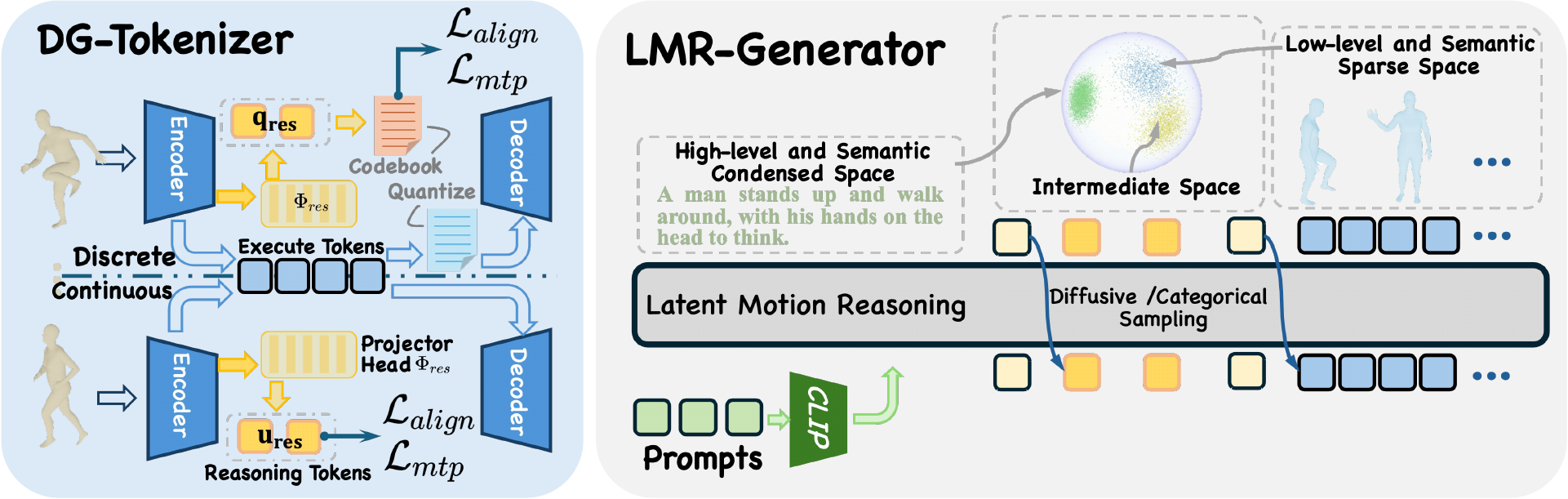}
    \caption{\textbf{Overview of the proposed Latent Motion Reasoning (LMR) framework}, organized in pipeline order from the DG-Tokenizer to the LMR-Generator. \textbf{(Left) Dual-Granularity (DG) Tokenizer}: We explicitly disentangle motion representations into two manifolds: a compressed Reasoning Latent ($\mathbf{q}_{res}$/$\mathbf{u}_{res}$, Yellow), which is aligned with text embeddings to capture high-level semantic intent via $\mathcal{L}_{align}$ and $\mathcal{L}_{mtp}$, and a high-frequency \textbf{Execution Latent} (Blue), which follows the baseline reconstruction (e.g., the T2M-GPT or MotionStreamer motion autoencoder); the panel shows the training of the reasoning branch, supervised by $\mathcal{L}_{align}$ and $\mathcal{L}_{mtp}$. \textbf{(Right) LMR-Generator}: We reformulate T2M as a hierarchical ``Think-then-Act'' generation process. Conditioned on the text prompt, the model first autoregressively synthesizes the coarse-grained reasoning tokens to establish the global motion topology (Thinking Phase). These tokens then serve as a stable semantic condition to guide the subsequent generation of fine-grained execution tokens (Acting Phase) via either Categorical or Diffusive sampling, a forward pass shared by training and inference. The intermediate latent manifold (top) is a conceptual illustration; see Fig.~\ref{fig:tsne_manifold} for the actual visualization.}
    \label{fig:pipeline}
\end{figure*}

%% file: Sections/4_new_experiments.tex
\section{Experiment}
\input{Tables/table_main_results}
\subsection{Experimental Setup}
\label{subsec:setup}
\noindent\textbf{Datasets.}
We evaluate our method on two standard motion-language benchmarks: HumanML3D~\cite{HumanML3D} and KIT-ML~\cite{KIT-ML}.
The HumanML3D dataset comprises 14,616 motions sourced from AMASS~\cite{AMASS} and HumanAct12~\cite{HumanAct12}, with each motion annotated by three text descriptions, totaling 44,970 descriptions.
The KIT-ML dataset provides a smaller-scale evaluation platform with 3,911 motions and 6,278 text descriptions.
All motion sequences are split into training (80\%), testing (15\%), and validation (5\%) sets.

\noindent\textbf{Metrics.} We adhere to the standard evaluation protocol proposed in T2M~\cite{T2M}, and report each metric together with the property it characterizes.
We treat \emph{physical fidelity}---the degree to which a generated motion is physically plausible as real human motion---as a two-level concept.
At the \emph{macro} level, Fr\'echet Inception Distance (\textbf{FID})~\cite{FID} measures the distributional divergence between synthesized and ground-truth motions, i.e.\ how closely the generation follows real human-motion statistics as a whole; where the distinction matters we write \textbf{rFID} (\emph{reconstruction FID}, computed on tokenizer reconstructions) and \textbf{gFID} (\emph{generation FID}, computed on text-conditioned generations).
At the \emph{micro} level, we additionally report dedicated kinematic metrics---Physical Foot Contact score (\textbf{PFC})~\cite{EDGE}, \textbf{Float} and \textbf{Foot-contact}~\cite{PhysiInter}---that capture local physical violations such as foot sliding, floating, and ground penetration, whose mathematical details are given in Appendix~\ref{app:metrics}.
Beyond physical fidelity, Retrieval Precision (\textbf{R-Precision}) (Top-1/2/3) and Multimodal Distance (\textbf{MM-Dist}) evaluate semantic alignment between text and motion; \textbf{Diversity} measures the variance of generated motions; Mean Per Joint Position Error (\textbf{MPJPE}) measures execution/reconstruction fidelity; and masked-keyword prediction Accuracy (\textbf{ACC}), the prediction accuracy of masked keywords, serves as a proxy for local reasoning alignment.
Appendix~\ref{app:metrics} gives the mathematical details of the physical-fidelity metrics, argues why this set is representative, and summarizes in Table~\ref{tab:metric_property} the full name, definition and certified property of every metric used in this paper.

\noindent\textbf{Implementation Details.}
We implement our framework in PyTorch and train on NVIDIA H200 GPUs. We instantiate LMR with two backbones to validate its generality.

\textit{Discrete Setting.}
We adopt the 263-dimensional joint position representation~\cite{HumanML3D} and VQ-VAE architecture from~\cite{T2M-GPT} with codebook size 512.
The execution encoder is first trained with batch size 256 and learning rate $2e{-4}$.
We then freeze it and train the reasoning branch for 200K iterations with $\lambda_{align}=0.5$ and $\lambda_{mtp}=0.1$.
The generator follows T2M-GPT with 18 Transformer layers and 1024 hidden dimensions, trained for 300K iterations with batch size 256.
During inference, we apply Classifier-Free Guidance with scale $s=2.0$.

\textit{Continuous Setting.}
We employ the 272-dimensional SMPL rotation representation~\cite{MotionStreamer}, which bypasses Inverse Kinematics post-processing.
The execution branch adopts a Causal Temporal AutoEncoder.
After training the execution branch for 2M iterations, we freeze it and train the reasoning branch for 200K iterations with learning rate $1e{-4}$ and identical loss weights with $\lambda_{align}=0.5$, $\lambda_{mtp}=0.1$.
The generator comprises a 12-layer Transformer (768 hidden dim), incorporated with a 9-layer MLP diffusion head.
Training proceeds for 300K iterations with batch size 256 and cosine learning rate schedule (peak $1e{-4}$).
During inference, we set the guidance scale $s=5.0$.

\textit{Reasoning Branch Architecture.}
The reasoning branch operates on the encoded motion representations from the execution branch encoder. Specifically, it employs a 2-layer causal Transformer with 4 attention heads and 512 hidden dimensions, with a feed-forward dimension of 1024, followed by a 4x temporal downsampling module. For the discrete setting, a vector quantizer with codebook size 512 discretizes these features into reasoning tokens. For the
continuous setting, a VAE encoder maps the features into a 16-dimensional latent space.
\input{NewFigures/Fig_quality_result}
\input{Tables/table_main_motionstreamer_results}

\textit{Masked Token Prediction.}
For $\mathcal{L}_{mtp}$, we employ a frozen BERT-base model with structured masking targeting action verbs, body parts, directional modifiers, and their compositions. The reasoning tokens are fed into a 2-layer cross-attention Transformer decoder with 4 heads, 768 hidden dimensions, and 1536 feed-forward dimensions to predict masked text tokens, encouraging fine-grained semantic learning. Dropout is applied during training.

\subsection{Comparison with State-of-the-Arts}

\noindent\textbf{Quantitative Results.}
Table~\ref{tab:sota} and Table~\ref{tab:motionstreamer_comparison} present a comprehensive comparison against state-of-the-art Diffusion, GPT, BERT, and Hybrid frameworks. Our analysis yields three key observations regarding the efficacy of Latent Motion Reasoning (LMR):

\textit{Architectural Decoupling vs. Component Engineering (Table~\ref{tab:sota}).}
On the discrete benchmark, LMR significantly outperforms its backbone T2M-GPT (FID 0.141$\longrightarrow$\textbf{0.040} on HumanML3D) and advanced GPT variants such as ParCo~\cite{ParCo} and the strong MOGO$^{\dagger}$~\cite{Mogo} (FID 0.040 vs.\ 0.064), which respectively scale codebook capacity and decompose body parts. It also surpasses the latest diffusion model FG-T2M++~\cite{FGT2Mpp} (FID 0.040 vs.\ 0.089), whose gains stem from heavy text-side engineering---LLM-based fine-grained parsing and hyperbolic syntactic embeddings. This indicates that the semantic bottleneck in autoregressive modeling is fundamental: closing it requires explicitly decoupling reasoning from execution, rather than scaling codebook capacity, decomposing body parts, or enriching the textual front-end.

\textit{Planning vs. Correction (Table~\ref{tab:sota}).}
Compared to BERT-based masked models (MMM, MoMask), LMR achieves superior diversity and semantic alignment without relying on ground-truth motion length—a critical limitation in MoMask. Furthermore, LMR outperforms BAMM, a hybrid method that generates then refines. This validates our core hypothesis: planning the topology before execution (``Think-then-Act'') is a more effective inductive bias than generating followed by post-hoc correction. The recent masked model Motion Anything~\cite{MotionAnything} attains stronger distributional and retrieval scores on HumanML3D (and a lower FID on KIT-ML); we attribute this to its multi-step bidirectional refinement, which sharpens distributional metrics at the cost of causality, streaming generation, a motion-length prior, and KV-cache--based efficient inference. LMR deliberately preserves these autoregressive properties while reaching comparable quality as a plug-and-play reasoning module on a vanilla AR backbone, and remains the best performer on KIT-ML across retrieval and MM-Dist.

\input{NewFigures/Fig_motionstreamer_quality}
\textit{Generalizability and Representation Sensitivity (Table~\ref{tab:motionstreamer_comparison}).}
LMR proves agnostic to the underlying representation. In the continuous setting, it improves the MotionStreamer~\cite{MotionStreamer} backbone to achieve the lowest FID of \textbf{9.937}. Notably, the performance leap is more pronounced in the discrete domain than the continuous one. We attribute this to the nature of the baselines: discrete quantization creates severe ``Semantic Sparsity,'' where LMR acts as a critical structural stabilizer. In contrast, continuous diffusion models already benefit from smoothing; here, LMR serves primarily to sharpen semantic alignment rather than rescue the model from structural collapse.

\input{Tables/table_physical_metrics}
\textit{Physical Fidelity (Table~\ref{tab:physical_metrics}).}
We further assess the micro-level physical fidelity defined in Sec.~\ref{subsec:setup} using the dedicated kinematic metrics PFC, Float, and Foot-contact. As shown in Table~\ref{tab:physical_metrics}, under the same single-inference protocol LMR consistently outperforms the baselines (MLD, MoMask, T2M-GPT) on all metrics and closely matches the Real reference, providing direct local physical evidence that complements its strong macro-level FID.

\noindent\textbf{Qualitative Results.}
Visualizations in Figure~\ref{fig:quality_result} and Figure~\ref{fig:motionstreamer_quality_result} highlight the critical advantage of our hierarchical generation process.

\textit{Complex Semantic Reasoning (Figure~\ref{fig:quality_result}).} While standard baselines (T2M-GPT, ParCo) suffer from semantic incompleteness—often ignoring concurrent actions—and hybrid methods (BAMM) struggle with attribute binding (e.g., confusing left/right limbs), LMR accurately resolves strict spatial and compositional constraints. By establishing a global motion topology first, it prevents the structural degradation observed in single-stage models.

\textit{Logic-Driven Constraints (Figure~\ref{fig:motionstreamer_quality_result}).} In continuous latent spaces, LMR successfully enforces logic-driven requirements such as repetition counting and directional changes, whereas MotionStreamer frequently drifts, failing to maintain the causal logic of the prompt.

\textit{Latent Interpretability (Figure~\ref{fig:Fig_pred_reasoning_mask}).} Finally, we validate the semantic density of our reasoning tokens via Masked Token Prediction. The model's ability to recover masked keywords solely from the reasoning latent proves that this manifold captures high-level \textit{semantic intent} rather than mere kinematic compression, effectively bridging the linguistic-physical gap.
\input{NewFigures/Fig_pred_reasoning_mask}

\noindent\textbf{User Study.} Please refer to Appendix~\ref{app:user} for more details.

\input{Tables/table_ablation_token_number}

\subsection{Ablation Study}
The following ablation studies are conducted on the discrete T2M-GPT backbone with the HumanML3D dataset.
In addition to the below experiments, Cross-backbone analyses and Impact of Classifier-Free Guidance  involving both discrete and continuous settings are provided in Appendix~\ref{app:ablate}.

\subsubsection{ Impact of Dual-Granularity Tokenizer (Table~\ref{tab:ablation_settings}).}
We investigate how execution and reasoning token scales affect reconstruction, semantic alignment, and generation.
For execution tokens, higher resolution yields better reconstruction: $1\times$ achieves MPJPE of 16.608 (row 1), while $1/4\times$ degrades to 30.294 (row 2). However, directly generating high-resolution motion is challenging—note that the
$1/4\times$ model attains a significantly better gFID than $1\times$. This trend is further corroborated by rows 4, 7, and 9. Moreover, since reconstruction quality upper-bounds generation performance, we opt not to downsample the execution tokens.
For reasoning tokens, which form a \emph{compressed, abstract} plan and are therefore downsampled relative to (rather than matched to) the execution length, fixing execution at $1\times$ and varying the reasoning scale reveals that $1/4\times$ is the knee point with gFID of 0.040: $1/2\times$ stays too close to the high-frequency execution stream and under-abstracts, whereas $1/8\times$ over-compresses and discards the ordering of sub-actions, which explains our choice of $T/4$.
Additionally, the Joint baseline using a single $1\times$ sequence fails at both reconstruction and generation due to conflicting objectives, while our Dual strategy achieves optimal balance by decoupling execution and reasoning manifolds. Performance further varies smoothly with the relative loss weighting $\lambda_{align}\!:\!\lambda_{mtp}$ and the reasoning codebook size, confirming robustness to these remaining hyper-parameters (see Appendix~\ref{app:hparam}).

\input{Tables/table_ablation_token_training}

\subsubsection{ Training Pipeline Analysis (Table~\ref{tab:ablation_token_training}).}
We evaluate four training strategies for our Dual-Granularity Tokenizer.

\emph{End-to-End joint} training yields suboptimal generation with gFID of 0.145, as conflicting gradients between reconstruction and semantic objectives degrade execution quality to rFID=0.133 and MPJPE=22.563.

\emph{Two-Stage Freeze} strategy we employed achieves optimal performance with gFID of 0.040 and MM-Dist of 2.895 by first establishing a high-fidelity execution substrate with rFID of 0.044 and MPJPE of 16.608, then freezing it during reasoning branch training to prevent representational drift.

\emph{Two-Stage Fine-tune} degrades execution quality to rFID of 0.101 despite marginally higher semantic alignment at ACC=0.549, resulting in inferior generation with gFID=0.109. This confirms the execution space should remain invariant after initialization.

\emph{Independent Networks} achieve identical reconstruction and the highest semantic alignment with R1=0.489 and ACC=0.556, yet generation degrades to gFID of 0.097. This stems from architectural disconnection: our reasoning tokens derive from execution encoder features, grounding semantic abstractions in the kinematic manifold. Independent encoders create disjoint representation spaces, introducing a semantic-kinematic gap that impairs generation. This validates our design—shared encoder backbone ensures representational coherence while frozen parameters prevent optimization interference.

\input{Tables/tabel_Different_Options_of_Semantic_Sequence}

\subsubsection{ Effectiveness of Guidance Strategies.}
Table~\ref{tab:ablation_strategies} validates the necessity of our learned reasoning tokens by comparing them against three alternative guidance paradigms.

\textit{Latent Representation vs. Off-the-Shelf Embeddings.}
Replacing our reasoning tokens with pre-trained TMR embeddings~\cite{TMR} degrades R-Precision to 0.472. This confirms that generic motion-language embeddings lack the specific semantic discriminability required for generation. Our reasoning tokens, optimized via joint global alignment and local masked prediction, encode significantly richer control signals.

\textit{Manifold Disentanglement vs. Temporal Coarsening.}
A naive Coarse-to-Fine autoregressive baseline (predicting downsampled tokens first) yields a poor FID of 0.154 due to error accumulation. This comparison isolates the source of our performance: success stems not merely from progressive generation, but from Manifold Disentanglement. Unlike coarse-to-fine approaches that operate entirely within the kinematic space, LMR establishes the trajectory in a separate, semantically aligned manifold before mapping to execution. This prevents the generation of ``compressed but empty'' kinematics that lack semantic grounding.

\textit{Motion-Aligned Reasoning vs. Linguistic Reasoning.}
Finally, we examine explicit linguistic guidance via full CLIP sequences~\cite{CLIP} and Textual Chain-of-Thought (Motion-R1 style~\cite{MotionR1}). While these methods improve semantic consistency (R-Precision 0.523 and 0.511), their generation quality remains suboptimal (FID 0.206 and 0.188).
This validates that fine-grained text can partially improve performance. However, symbolic text remains too sparse to effectively guide high-frequency motion synthesis, illustrating the ineffability of physical dynamics.
Our approach achieves the best balance (FID 0.040, R-Precision 0.537) by conducting reasoning in a learned, motion-aligned latent substrate—effectively bridging the gap between abstract intent and physical execution.

\input{Tables/table_placebo}

\subsubsection{ Causal Analysis of Latent Reasoning (Table~\ref{tab:placebo}).}
Having shown that the learned reasoning tokens outperform alternative guidance signals, we next verify that this advantage truly comes from the \emph{content} they carry rather than from extra tokens, parameters, or prefix capacity. We conduct a reasoning-token placebo experiment: keeping the architecture, prefix length, parameter count, and token budget fixed, we perturb only the \emph{content} of the reasoning tokens before regenerating the execution stream autoregressively. We observe two distinct perturbation regimes. (i) \emph{Content perturbed but still valid} (Random/Shuffle): FID stays essentially unchanged and even slightly improves (0.040$\rightarrow$0.031/0.038), yet retrieval drops markedly (Top-1 0.537$\rightarrow$0.493/0.497, MM-Dist 2.895$\rightarrow$3.110/3.073). The slight FID drop is expected: as the text constraint weakens, the model is no longer pinned to the specific (often rarer) action the prompt requests and instead drifts toward the \emph{average / typical} actions of the dataset, which are by construction closest to the overall data distribution; a distribution-level metric such as FID therefore does not penalize them and can even go slightly lower. A good FID here is thus \emph{not} evidence that the reasoning content is preserved---the retrieval drop reveals that the motion has quietly decoupled from the text while remaining a generic, distribution-typical sequence. This confirms that the reasoning stream encodes a \emph{specific, text-aligned plan}, not a generic motion prior. (ii) \emph{Information fully erased} (Constant/Zero): both axes collapse, with FID worsening $\sim$4.7$\times$ (0.040$\rightarrow$0.183/0.184) and Top-1 falling to 0.460/0.446. As the only variable is reasoning-token content, the improvement is causally attributable to the reasoning content itself.

\input{Tables/table_ablation_semantic_learning}

\subsubsection{ Reasoning Loss Functions (Table~\ref{tab:ablation_semantic_learning}).}
Given that the reasoning content is what drives the gain, we finally examine the objectives that instill this content---$\mathcal{L}_{align}$ for global alignment and $\mathcal{L}_{mtp}$ for local alignment.
$\mathcal{L}_{align}$ optimizes global R-Precision but neglects local details with an ACC of 0.000.
$\mathcal{L}_{mtp}$ captures fine-grained semantics yielding an ACC of 0.534 but loses global coherence.
Combining both in our approach provides the most effective synergy, ensuring the generated plan is both globally consistent and locally precise.

\input{Tables/table_effectiveness_off-the-shelf}

\subsubsection{ Generalizability of Dual-Granularity Strategy (Table~\ref{tab:comparison_tokenizer}).}
To verify the universality of LMR, we compare tokenization strategies on both discrete (T2M-GPT) and continuous (MotionStreamer) backbones in Table~\ref{tab:comparison_tokenizer}. Against the best single-granularity baseline, our Dual strategy improves FID from 0.137 to \textbf{0.040} in discrete space and from 11.790 to \textbf{9.937} in continuous space, confirming that the reasoning layer bridges the semantic-kinematic gap regardless of backbone. The larger discrete gain reflects its severe semantic sparsity, where reasoning acts as a critical stabilizer; the diffusion-smoothed continuous space instead benefits mainly from sharpened semantic alignment.

%% file: Tables/table_main_results.tex
\begin{table*}[thb]
    \centering
    \caption{Comparison with diffusion-, BERT-, and GPT-type models of text-conditional motion synthesis on the HumanML3D and KIT-ML test set. $\pm$ indicates a 95\% confidence interval. Among the GPT- and BERT/mask-type methods, we indicate the best result in \textbf{bold} face, and the second best in \underline{underscore}. $^{\dagger}$For a fair single-pass comparison, MOGO is reported without its Text-Condition-Alignment (TCA) post-hoc module, consistent with all other methods.}
    \renewcommand{\arraystretch}{1.15}{
    \scalebox{0.9}{
    \begin{tabular}{l l c c c c c c}
    \toprule
    \multirow{2}{*}{Datasets} & \multirow{2}{*}{Methods}  & \multicolumn{3}{c}{R Precision$\uparrow$} & \multirow{2}{*}{FID$\downarrow$} & \multirow{2}{*}{MultiModal Dist$\downarrow$} & \multirow{2}{*}{Diversity $\rightarrow$}\\

    \cmidrule(lr){3-5}
       ~& ~ & Top 1 & Top 2 & Top 3 \\

    \midrule
    \multirow{15}{*}{\makecell[c]{Human\\ML3D}}
        ~& Ground Truth & \et{0.511}{.003} & \et{0.703}{.003} & \et{0.797}{.002} & \et{0.002}{.000} & \et{2.974}{.008} & \et{9.503}{.065}  \\
        \cline{2-8}
        ~& MDM~\cite{MDM} & - & - & \et{0.611}{.007} & \et{0.544}{.044} & \et{5.566}{.027} & \et{9.559}{.086}  \\
        ~ & MLD~\cite{MLD} & \et{0.481}{.003} & \et{0.673}{.003} & \et{0.772}{.002} & \et{0.473}{.013} & \et{3.196}{.010} & \et{9.724}{.082}  \\
        ~ & MotionDiffuse~\cite{MotionDiffuse} & \et{0.491}{.001} & \et{0.681}{.001} & \et{0.782}{.001} & \et{0.630}{.001} & \et{3.113}{.001}  & \et{9.410}{.049}  \\
        ~ & FG-T2M++~\cite{FGT2Mpp} & \et{0.513}{.002} & \et{0.702}{.002} & \et{0.801}{.003} & \et{0.089}{.004} & \et{2.925}{.007} & \et{9.223}{.114}  \\
        \cline{2-8}

        ~& MMM~\cite{MMM} &\et{0.504}{.003} & \et{0.696}{.003} & \et{0.794}{.004} & \et{0.080}{.004} & \et{2.998}{.007} & \et{9.411}{.058}\\

        ~ & MoMask~\cite{MoMask} & \et{0.521}{.002} & \et{0.713}{.002} & \et{0.807}{.002} & \et{0.045}{.002} & \et{2.958}{.008} & -\\

        ~ & BAMM~\cite{BAMM} & \et{0.525}{.002} & \et{0.720}{.003} & \ets{0.814}{.003} & \et{0.055}{.002} & \et{2.919}{.008} & \et{9.717}{.089} \\

        ~ & Motion Anything~\cite{MotionAnything} & \etb{0.546}{.003} & \etb{0.735}{.002} & \etb{0.829}{.002} & \etb{0.028}{.005} & \etb{2.859}{.010} & \etb{9.521}{.083} \\

        \cline{2-8}

        ~ & MotionGPT~\cite{MotionGPT} & \et{0.492}{.003} & \et{0.681}{.003} & \et{0.778}{.002} & \et{0.232}{.008} & \et{3.096}{.009} & \ets{9.528}{.071} \\

        ~ & ParCo~\cite{ParCo} & \et{0.515}{.003} & \et{0.706}{.003} & \et{0.801}{.002} & \et{0.109}{.005}  & \et{2.927}{.008} & \et{9.576}{.088} \\

        ~ & MOGO$^{\dagger}$~\cite{Mogo} & \et{0.515}{.003} & \et{0.709}{.003} & \et{0.801}{.003} & \et{0.064}{.002} & \et{2.951}{.008} & - \\

        ~ & T2M-GPT~\cite{T2M-GPT} & \et{0.492}{.003} & \et{0.679}{.002} & \et{0.775}{.002} & \et{0.141}{.005} & \et{3.121}{.009}  & \et{9.761}{.151}  \\

        \cline{2-8}
        \rowcolor[HTML]{EFEFEF}
        ~ & LMR(Ours) & \ets{0.537}{.005} & \ets{0.721}{.004} & \et{0.810}{.005} & \ets{0.040}{.002} & \ets{2.895}{.017} & \et{9.668}{.077}\\

    \midrule
    \multirow{15}{*}{\makecell[c]{KIT-\\ML}}
        ~& Ground Truth & \et{0.424}{.005} & \et{0.649}{.006} & \et{0.779}{.006} & \et{0.031}{.004} & \et{2.788}{.012} & \et{11.080}{.097}  \\
        \cline{2-8}
        ~& MDM~\cite{MDM} & - & - & \et{0.396}{.004} & \et{0.497}{.021} & \et{9.191}{.022} & \et{10.847}{.109}  \\

        ~& MLD~\cite{MLD} & \et{0.390}{.008} & \et{0.609}{.008} & \et{0.734}{.007} & \et{0.404}{.027} & \et{3.204}{.027} & \et{10.80}{.117}  \\

        ~& MotionDiffuse~\cite{MotionDiffuse} & \et{0.417}{.004} & \et{0.621}{.004} & \et{0.739}{.004} & \et{1.954}{.062} & \et{2.958}{.005}  & \et{11.10}{.143}  \\

        ~ & FG-T2M++~\cite{FGT2Mpp} & \et{0.442}{.006} & \et{0.657}{.005} & \et{0.781}{.004} & \et{0.135}{.004} & \et{2.696}{.011} & \et{10.99}{.105}  \\

        \cline{2-8}

        ~& MMM~\cite{MMM} & \et{0.381}{.005} & \et{0.590}{.006} & \et{0.718}{.005} & \et{0.429}{.019} & \et{3.146}{.019} & \et{10.633}{.097} \\

        ~& MoMask~\cite{MoMask} & \et{0.433}{.007} & \et{0.656}{.005} & \et{0.781}{.005} & \et{0.204}{.011} & \et{2.779}{.022} & - \\

        ~ & BAMM~\cite{BAMM} & \et{0.438}{.009} & \et{0.661}{.009} & \et{0.788}{.005} & \et{0.183}{.013} & \et{2.723}{.026} & \ets{11.008}{.094} \\

        ~ & Motion Anything~\cite{MotionAnything} & \ets{0.449}{.007} & \ets{0.678}{.004} & \ets{0.802}{.006} & \etb{0.131}{.003} & \ets{2.705}{.024} & \et{10.94}{.098} \\

        \cline{2-8}

        ~& MotionGPT~\cite{MotionGPT} & \et{0.366}{.005} & \et{0.558}{.004} & \et{0.680}{.005} & \et{0.510}{.004} & \et{3.527}{.021} & \et{10.35}{.084} \\

        ~& ParCo~\cite{ParCo} & \et{0.430}{.004} & \et{0.649}{.007} & \et{0.772}{.008} & \et{0.453}{.027} & \et{2.820}{.028} & \et{10.95}{.094} \\

        ~& MOGO$^{\dagger}$~\cite{Mogo} & \et{0.420}{.007} & \et{0.634}{.007} & \et{0.754}{.007} & \et{0.313}{.016} & \et{2.957}{.029} & - \\

        ~& T2M-GPT~\cite{T2M-GPT} & \et{0.416}{.006} & \et{0.627}{.006} & \et{0.745}{.006} & \et{0.514}{.029} & \et{3.007}{.023}  & \et{10.921}{.108}  \\
        \cline{2-8}
        \rowcolor[HTML]{EFEFEF}
        ~& LMR(Ours) &\etb{0.483}{.012} & \etb{0.701}{.007} & \etb{0.811}{.006} & \ets{0.181}{.006} & \etb{2.636}{.014} & \etb{11.032}{.128}\\
    \bottomrule
    \end{tabular}
    }
    }
    \label{tab:sota}

\end{table*}

%% file: NewFigures/Fig_quality_result.tex
\begin{figure*}
    \centering
    \includegraphics[width=\linewidth]{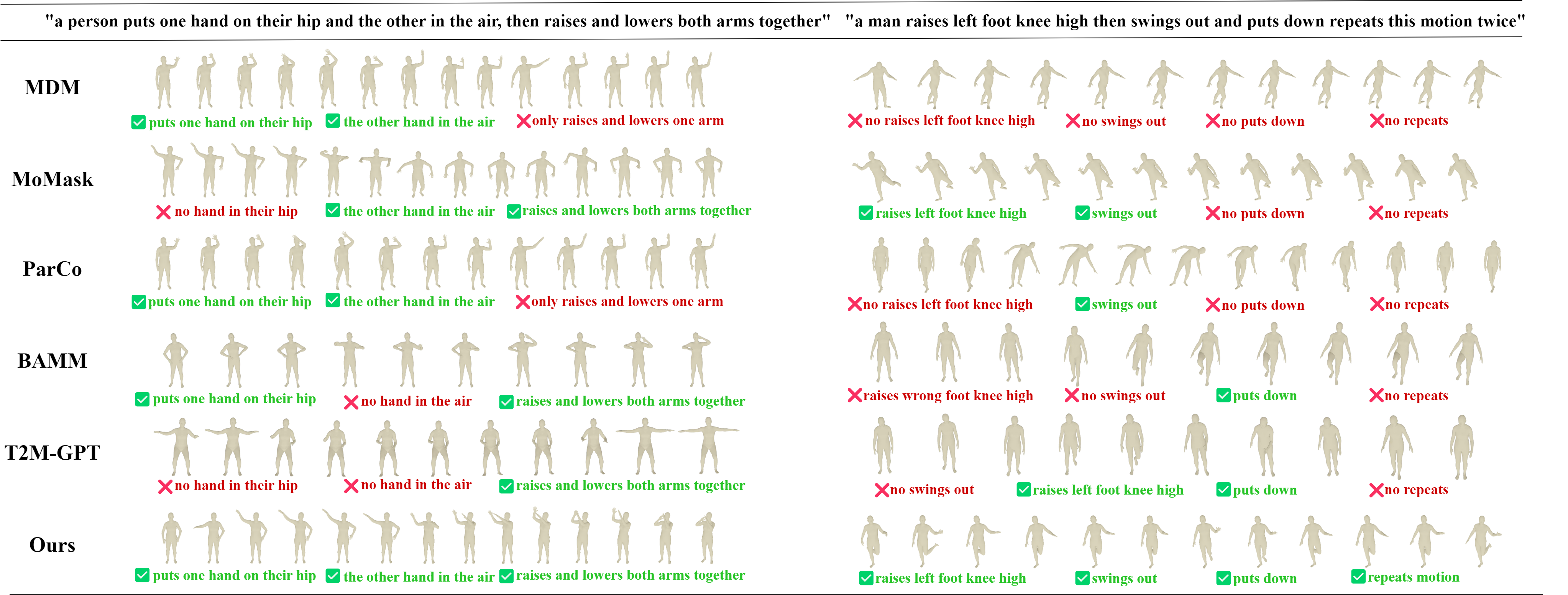}
    \caption{Qualitative comparison with state-of-the-art methods under the discrete motion representation setting.}
     \label{fig:quality_result}
\end{figure*}

%% file: Tables/table_main_motionstreamer_results.tex
\begin{table}[t]
    \centering
    \caption{Quantitative comparison on HumanML3D test set under the \textbf{continuous motion representation setting}~\cite{MotionStreamer}. Real motion serves as an upper-bound reference. Lower FID and MultiModal Dist are better; higher R-Precision and Diversity are better. Among \emph{generated} methods, \textbf{bold} denotes best, \underline{underline} denotes second best.}
    \resizebox{0.5\textwidth}{!}{
    \begin{tabular}{l c c c c c c}
    \toprule
    \multirow{2}{*}{Methods} & \multicolumn{3}{c}{R Precision$\uparrow$} & \multirow{2}{*}{FID$\downarrow$} & \multirow{2}{*}{MultiModal Dist$\downarrow$} & \multirow{2}{*}{Diversity$\rightarrow$} \\
    \cmidrule(lr){2-4}
     & Top 1 & Top 2 & Top 3 \\
    \midrule
    Real motion & 0.702 & 0.864 & 0.914 & 0.002 & 15.151 & 27.492 \\
    \midrule
    T2M-GPT~\cite{T2M-GPT} & 0.606 & 0.774 & 0.838 & 12.475 & 16.812 & 27.275 \\
    MotionGPT~\cite{MotionGPT} & 0.456 & 0.598 & 0.628 & 14.375 & 17.892 & 27.114 \\
    Momask~\cite{MoMask} & 0.621 & 0.784 & 0.846 & 12.232 & 16.138 & 27.127 \\
    AttT2M~\cite{AttT2M} & 0.592 & 0.765 & 0.834 & 15.428 & \textbf{15.726} & 26.674 \\
    MotionStreamer~\cite{MotionStreamer} & \underline{0.631} & \underline{0.802} & \underline{0.859} & \underline{11.790} & {16.081} & \underline{27.284} \\
    \rowcolor{gray!15}
    LMR(Ours) & \textbf{0.644} & \textbf{0.812} & \textbf{0.869} & \textbf{9.937} & \underline{15.968} & \textbf{27.373} \\
    \bottomrule
    \end{tabular}
    }
    \label{tab:motionstreamer_comparison}
\end{table}

%% file: NewFigures/Fig_motionstreamer_quality.tex
\begin{figure}
    \centering
    \includegraphics[width=\linewidth]{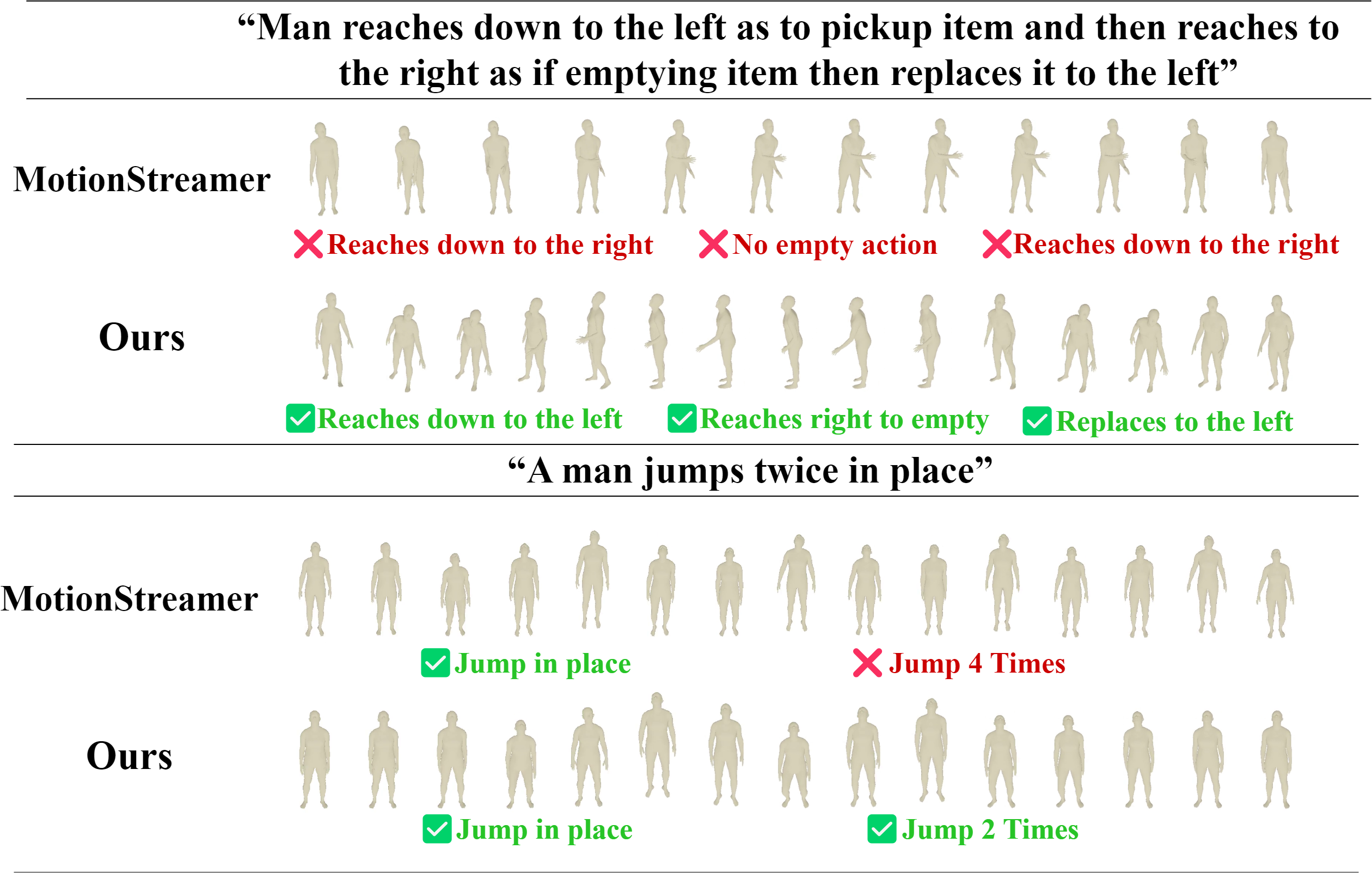}
    \caption{Qualitative comparison with MotionStreamer~\cite{MotionStreamer} under the continuous motion representation setting.}
     \label{fig:motionstreamer_quality_result}
\end{figure}

%% file: Tables/table_physical_metrics.tex
\begin{table}[h]
\centering
\caption{Physical-fidelity metrics (lower is better), with Real motion as the reference. All methods use official weights and the same single-inference protocol.}
\label{tab:physical_metrics}
\setlength{\tabcolsep}{4pt}
\footnotesize
\begin{tabular}{lcccc}
\toprule
Method & FID $\downarrow$ & PFC $\downarrow$ & Float $\downarrow$ & Foot-contact $\downarrow$ \\
\midrule
Real             & 0.002 & 0.425 & 0.0212 & 0.0630 \\
MLD              & 0.473 & 0.802 & 0.0304 & 0.0640 \\
MoMask           & 0.045 & 0.685 & 0.0278 & 0.0683 \\
T2M-GPT          & 0.141 & 0.545 & 0.0231 & 0.0663 \\
\rowcolor[HTML]{EFEFEF}
\textbf{LMR}     & \textbf{0.040} & \textbf{0.523} & \textbf{0.0214} & \textbf{0.0632} \\
\bottomrule
\end{tabular}
\end{table}

%% file: NewFigures/Fig_pred_reasoning_mask.tex
\begin{figure}
    \centering
    \includegraphics[width=\linewidth]{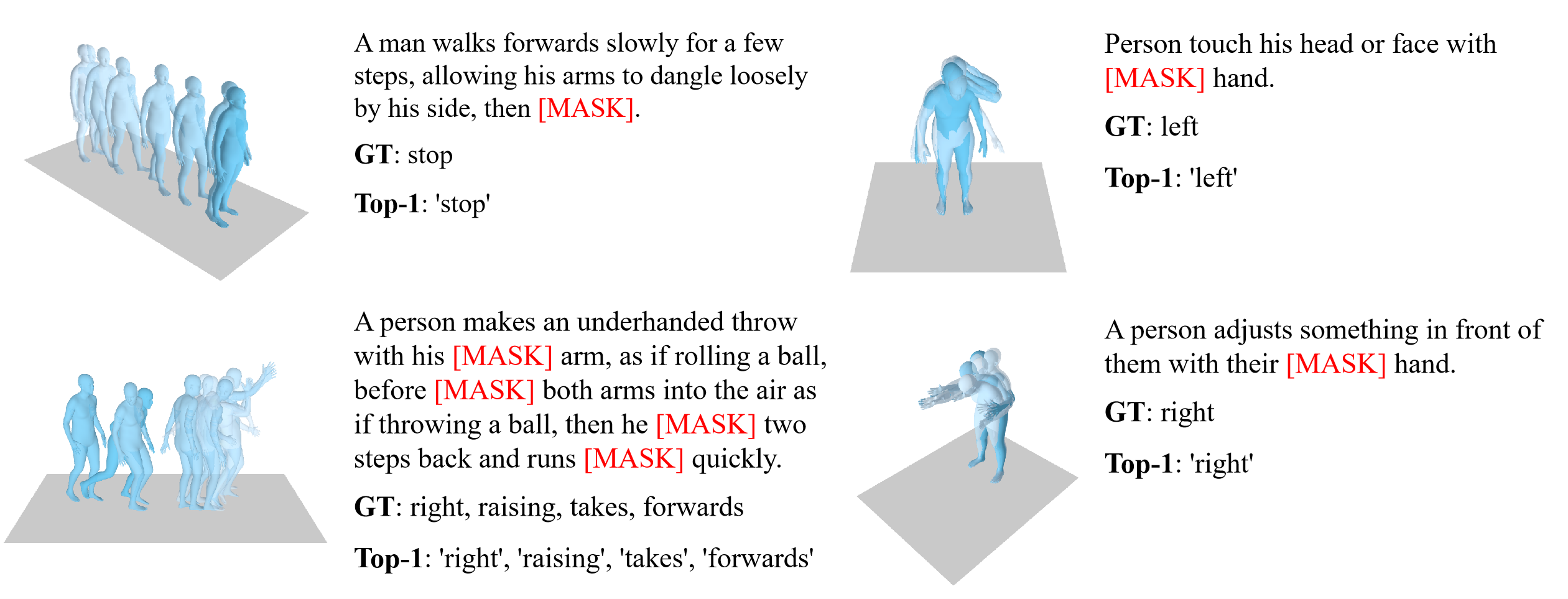}
    \caption{Visualization of the Masked Token Prediction (MTP) capability. The model predicts masked keywords (marked in red) such as body parts (`left', `right') and actions (`stop', `raising') using the learned Reasoning Latents. The high accuracy of Top-1 predictions demonstrates that our reasoning module effectively captures fine-grained semantic-kinematic alignment.}
     \label{fig:Fig_pred_reasoning_mask}
\end{figure}

%% file: Tables/table_ablation_token_number.tex
\begin{table}[htbp]
    \centering
    \caption{Ablation study of different token configurations. Exec. and Reas. denote the token sequence length ratio ($T/T_m$). The Joint setting implies a single sequence handling both tasks.}
    \label{tab:ablation_settings}
    \setlength{\tabcolsep}{2.5pt} 
    \resizebox{.5\textwidth}{!}{
    \begin{tabular}{l cc c cc c cc}
    \toprule
    \multirow{2}{*}{Method} & \multicolumn{2}{c}{Token Scale} & \multicolumn{2}{c}{Execution Quality} & \multicolumn{2}{c}{Reasoning Alignment} & \multicolumn{2}{c}{Generation}\\
    \cmidrule(lr){2-3} \cmidrule(lr){4-5} \cmidrule(lr){6-7} \cmidrule(lr){8-9}
     & Exec. & Reas. & rFID$\downarrow$ & MPJPE$\downarrow$ & R1$\uparrow$ & ACC$\uparrow$ & gFID$\downarrow$ & MM-Dist$\downarrow$\\
    \midrule
    \multirow{3}{*}{Baseline}
      & $1/4$ & -- & 0.095 & 30.294 & -- & -- & 0.137 & 3.225 \\
      & $1$ & -- & 0.044 & 16.608 & --    & --    & 0.242 & 3.148 \\ 
      & \multicolumn{2}{c}{$1$ (Joint)} & 0.083 & 22.512 & 0.429 & 0.463 & 0.354 & 3.253 \\
    \midrule
    \multirow{6}{*}{Ours}
      & $1$ & $1/2$  & 0.044 & 16.608 & 0.466 & 0.536 & 0.212 & 3.250 \\
      & \textbf{$\mathbf{1}$} & \textbf{$\mathbf{1/4}$} & \textbf{0.044} & \textbf{16.608} & \textbf{0.474} & \textbf{0.538} & \textbf{0.040} & \textbf{2.895} \\
      & $1$ & $1/8$  & 0.044 & 16.608 & 0.462 & 0.533 & 0.062 & 3.145 \\
      & $1/2$ & $1/8$ & 0.094 & 24.790 & 0.468 & 0.526 & 0.069 & 3.182 \\
      & $1/2$ & $1/16$& 0.094 & 24.790 & 0.455 & 0.519 & 0.076 & 2.969 \\
      & $1/4$ & $1/16$& 0.095 & 30.294 & 0.463 & 0.533 & 0.105 & 3.092 \\
    \bottomrule
    \end{tabular}
    }
\end{table}

%% file: Tables/table_ablation_token_training.tex
\begin{table}[htbp]
    \centering
    \caption{Ablation study on the training strategies for the dual-branch tokenizer. End-to-End denotes joint training; Two-Stage implies training the execution branch first, followed by the reasoning branch (with freezing or fine-tuning); Independent uses two separate networks.}
    \label{tab:ablation_token_training}
     \setlength{\tabcolsep}{2pt} 
     \scriptsize
     \resizebox{\columnwidth}{!}{%
    \begin{tabular}{l cc cc cc}
    \toprule
    \multirow{2}{*}{Training Strategy}  & \multicolumn{2}{c}{Execution Quality} & \multicolumn{2}{c}{Reasoning Alignment} & \multicolumn{2}{c}{Generation}\\
     \cmidrule(lr){2-3} \cmidrule(lr){4-5} \cmidrule(lr){6-7}
         & rFID$\downarrow$ & MPJPE$\downarrow$ & R1$\uparrow$ & ACC$\uparrow$ & gFID$\downarrow$ & MM-Dist$\downarrow$  \\
     \midrule

    End-to-End (Joint)    & 0.133 & 22.563 & 0.499 & 0.519 & 0.145 & 3.174\\

    \rowcolor[HTML]{EFEFEF}
    \textbf{Two-Stage (Freeze)} & \textbf{0.044} & \textbf{16.608} & 0.474 & 0.538 & \textbf{0.040} & \textbf{2.895}\\

    Two-Stage (Finetune)  & 0.101 & 20.276 & 0.481 & 0.549 & 0.109 & 3.077\\
    Independent Networks  & 0.044 & 16.608 & \textbf{0.489} & \textbf{0.556} & 0.097 & 3.134\\

    \bottomrule
    \end{tabular}
    }
\end{table}

%% file: Tables/tabel_Different_Options_of_Semantic_Sequence.tex
\begin{table}[t]
    \centering
    \caption{Ablation study on alternative guidance strategies. All variants adopt the high-resolution execution tokens ($1\times$). We compare five guidance schemes: off-the-shelf TMR embeddings~\cite{TMR}, coarse-to-fine autoregressive generation, extended CLIP token sequences~\cite{CLIP}, explicit language Chain-of-Thought, and our learned reasoning tokens.}
    \label{tab:ablation_strategies}
    \setlength{\tabcolsep}{4pt}
    \resizebox{0.48\textwidth}{!}{
    \begin{tabular}{l c ccc c c}
    \toprule
    \multirow{2}{*}{Guidance Strategy} & \multirow{2}{*}{FID$\downarrow$} & \multicolumn{3}{c}{R-Precision $\uparrow$} & \multirow{2}{*}{MM-Dist$\downarrow$} & \multirow{2}{*}{Div$\rightarrow$} \\
    \cmidrule(lr){3-5}
     & & Top-1 & Top-2 & Top-3 & & \\
    \midrule
    Baseline (w/o guidance) & 0.242 & 0.497 & 0.684 & 0.778 & 3.148 & 9.914 \\
    \midrule
    + TMR Embeddings & 0.120 & 0.472 & 0.666 & 0.759 & 3.210 & \textbf{9.965} \\
    + Coarse-to-Fine AR & 0.154 & 0.488 & 0.679 & 0.779 & 3.097 & 9.692 \\
    + CLIP Token ($L$=77) & 0.206 & 0.523 & 0.718 & \textbf{0.812} & 2.963 & 9.719 \\
    + Explicit Language CoT & 0.188 & 0.511 & 0.702 & 0.795 & 3.021 & 9.752 \\
    \midrule
    \textbf{+ Reasoning Tokens (Ours)} & \textbf{0.040} & \textbf{0.537} & \textbf{0.721} & 0.810 & \textbf{2.895} & 9.668 \\
    \bottomrule
    \end{tabular}
    }
    \vspace{-4mm}
\end{table}

%% file: Tables/table_placebo.tex
\begin{table}[h]
\centering
\caption{Reasoning-token placebo experiment on HumanML3D, perturbing only the reasoning-token content. \textbf{Normal}: original tokens; \textbf{Random}: random valid codes; \textbf{Shuffle}: order permuted; \textbf{Constant}: all codes set to $0$; \textbf{Zero}: embedding zeroed out.}
\label{tab:placebo}
\setlength{\tabcolsep}{4pt}
\footnotesize
\begin{tabular}{lccc}
\toprule
Mode & FID $\downarrow$ & Top1 $\uparrow$ & MM-Dist $\downarrow$ \\
\midrule
\rowcolor[HTML]{EFEFEF}
Normal   & \textbf{0.040} & \textbf{0.537} & \textbf{2.895} \\
Random   & 0.031 & 0.493 & 3.110 \\
Shuffle  & 0.038 & 0.497 & 3.073 \\
Constant & 0.183 & 0.460 & 3.349 \\
Zero     & 0.184 & 0.446 & 3.351 \\
\bottomrule
\end{tabular}
\end{table}

%% file: Tables/table_ablation_semantic_learning.tex
\begin{table}[t]
    \centering
    \caption{Ablation study on the reasoning loss functions. $\mathcal{L}_{align}$ focuses on aligning the reasoning tokens with the global text embedding, while $\mathcal{L}_{mtp}$ enforces fine-grained semantic understanding via masked text prediction. The combination achieves the best balance.}
    \label{tab:ablation_semantic_learning}
    \setlength{\tabcolsep}{3.8pt}
    \begin{tabular}{l cc c cc}
    \toprule
    \multirow{2}{*}{Reasoning Objective} & \multicolumn{2}{c}{Reasoning Alignment} & & \multicolumn{2}{c}{Generation Quality}\\
    \cmidrule(lr){2-3} \cmidrule(lr){5-6}
     & R1$\uparrow$ & ACC$\uparrow$& & FID$\downarrow$ & MM-Dist$\downarrow$ \\
    \midrule

    $\mathcal{L}_{align}$ & \textbf{0.482} & 0.000 & & 0.136 & 3.189\\
    $\mathcal{L}_{mtp}$ & 0.036 & 0.534 & & 0.120 & 3.161\\

    \rowcolor[HTML]{EFEFEF}
    \textbf{$\mathcal{L}_{align} + \mathcal{L}_{mtp}$ (Ours)} & 0.474 & \textbf{0.538} & & \textbf{0.040} & \textbf{2.895}\\

    \bottomrule
    \end{tabular}

\end{table}

%% file: Tables/table_effectiveness_off-the-shelf.tex
\begin{table}[htbp]
    \centering
    \caption{Comparison with different tokenizer strategies in Continuous and Discrete spaces. $\mathcal{D}$ denotes the temporal downsampling rate (e.g., $2\times$ implies sequence length $T/2$). 'Dual' indicates our proposed dual-granularity strategy.}
    \label{tab:comparison_tokenizer}
    \setlength{\tabcolsep}{3.2pt} 
    \resizebox{0.5\textwidth}{!}{
    \begin{tabular}{l c c ccc c c}
    \toprule
    \multirow{2}{*}{Method} & \multirow{2}{*}{$\mathcal{D}$} & \multirow{2}{*}{FID$\downarrow$} & \multicolumn{3}{c}{R-Precision $\uparrow$} & \multirow{2}{*}{MM-Dist$\downarrow$} & \multirow{2}{*}{Div$\rightarrow$} \\
    \cmidrule(lr){4-6}
     & & & Top-1 & Top-2 & Top-3 & & \\
    \midrule

    \multicolumn{8}{l}{\textit{\textbf{Continuous Latent Space}}} \\
    MotionStreamer & $2\times$ & 21.836 & 0.449 & 0.606 & 0.684 & 18.524 & 27.008 \\
    MotionStreamer & $4\times$ & 11.790 & 0.631 & 0.802 & 0.859 & 16.081 & 27.284 \\

    \rowcolor[HTML]{EFEFEF}
    \textbf{Ours} & \textbf{Dual} & \textbf{9.937} & \textbf{0.644} & \textbf{0.812} & \textbf{0.869} & \textbf{15.968} & \textbf{27.373} \\

    \midrule

    \multicolumn{8}{l}{\textit{\textbf{Discrete Latent Space}}} \\
    T2M-GPT & $1\times$ & 0.242 & 0.497 & 0.684 & 0.778 & 3.148 & \textbf{9.914} \\
    T2M-GPT & $4\times$ & 0.137 & 0.482 & 0.667 & 0.763 & 3.359 & 9.758 \\
    \rowcolor[HTML]{EFEFEF}
    \textbf{Ours} & \textbf{Dual} & \textbf{0.040} & \textbf{0.537} & \textbf{0.721} & \textbf{0.810} & \textbf{2.895} & 9.668 \\

    \bottomrule
    \end{tabular}
    }
\end{table}

%% file: Sections/5_conclusion.tex
\section{Conclusion}
This work redefines Text-to-Motion generation by identifying and resolving the \textbf{Semantic-Kinematic Impedance Mismatch}—the fundamental friction between abstract linguistic intent and continuous physical dynamics. We introduce \textbf{Latent Motion Reasoning (LMR)}, a framework that shifts the field from flat sequence-to-sequence translation to a hierarchical ``Think-then-Act'' paradigm. By decoupling global motion planning from precise kinematic execution via a \textbf{Dual-Granularity Tokenizer}, LMR consistently and substantially improves both discrete and continuous backbones, reaching performance on par with highly customized state-of-the-art methods while retaining the causality and efficiency of autoregressive generation. Crucially, our findings challenge the utility of explicit linguistic reasoning for physical tasks, demonstrating that the optimal substrate for motion planning is not natural language, but a learned, motion-aligned latent space. LMR represents a pivotal step toward \textbf{Cognitive Motion Generation}, moving beyond reflex-based synthesis to systems that possess a genuine, hierarchical understanding of physical behavior.

%% file: Sections/appendix.tex
\section{Evaluation Metrics: Definitions and Rationale}
\label{app:metrics}

This appendix makes the evaluation protocol of Sec.~\ref{subsec:setup} self-contained. We first give the mathematical details of the metrics that certify \emph{physical fidelity}, then argue why this particular set is representative, and finally summarize in Table~\ref{tab:metric_property} the full name, an informal definition and the certified property of every metric reported anywhere in this paper.

\noindent \textbf{Mathematical details of the physical-fidelity metrics.}

\noindent \emph{FID.} Let $\mathcal{F}(\cdot)$ be the motion feature extractor of Guo~et~al.~\cite{T2M}: a bidirectional GRU trained jointly with a text encoder under a contrastive text--motion matching objective, so that its feature space is learned from real human motion. For real motions $\{\mathbf{x}_i\}_{i=1}^{N}$ and generated motions $\{\hat{\mathbf{x}}_i\}_{i=1}^{N}$ we fit a Gaussian to each feature set,
\begin{equation}
\begin{aligned}
    \mu_r&=\frac{1}{N}\sum\nolimits_{i=1}^{N}\mathcal{F}(\mathbf{x}_i),\\
    \Sigma_r&=\frac{1}{N-1}\sum\nolimits_{i=1}^{N}\big(\mathcal{F}(\mathbf{x}_i)-\mu_r\big)\big(\mathcal{F}(\mathbf{x}_i)-\mu_r\big)^{\!\top},
\end{aligned}
\end{equation}
\noindent with $\mu_g,\Sigma_g$ defined identically from $\mathcal{F}(\hat{\mathbf{x}}_i)$, and report the Fr\'echet distance between the two~\cite{FID},
\begin{equation}
    \mathrm{FID}=\|\mu_g-\mu_r\|_2^2+\mathrm{Tr}\!\left(\Sigma_g+\Sigma_r-2\left(\Sigma_g\Sigma_r\right)^{1/2}\right),
\end{equation}
\noindent where $\mathrm{Tr}(\cdot)$ is the trace of a matrix, i.e.\ the sum of its diagonal entries. We stress one point that the historical name obscures: the FID reported in text-to-motion involves no Inception image network. Only the name is inherited; what is compared are distributions of human-motion features.

\noindent \emph{PFC (Physical Foot Contact score)}~\cite{EDGE}. EDGE~\cite{EDGE} starts from two observations: on the horizontal plane, any centre-of-mass (COM) acceleration must come from static foot--ground contact, and on the vertical axis the same holds for any \emph{upward} acceleration. PFC therefore penalises frames in which the COM accelerates while \emph{both} feet are in motion,
\begin{equation}
    s_i=\left\|\bar{\mathbf{a}}^{\mathrm{COM}}_{i}\right\|\cdot\left\|\mathbf{v}^{\mathrm{L}}_{i}\right\|\cdot\left\|\mathbf{v}^{\mathrm{R}}_{i}\right\|,
    \quad
    \mathrm{PFC}=\frac{\sum_{i=1}^{N}s_i}{N\max_{j}\left\|\bar{\mathbf{a}}^{\mathrm{COM}}_{j}\right\|},
\end{equation}
\noindent with $\bar{\mathbf{a}}^{\mathrm{COM}}_{i}=(a_{i,x},\,a_{i,y},\,\max(a_{i,z},0))^{\!\top}$: the horizontal components are kept as they are and the vertical one is clamped at zero, so that only upward acceleration is charged against foot contact. Following the official EDGE implementation, we approximate the centre of mass by the root joint, and take $\mathbf{v}^{\mathrm{L}}_{i}$ and $\mathbf{v}^{\mathrm{R}}_{i}$ to be the per-side minimum, over the ankle and toe joints, of the \emph{horizontal} inter-frame displacement, so that a side counts as planted when either of its two joints is stationary. A physically consistent motion drives at least one of the three factors to zero in every frame.

\noindent \emph{Float and Foot-contact}~\cite{PhysiInter}. These are defined in~\cite{PhysiInter} relative to the ground plane: \emph{Float} is the average height of the lowest point of the human mesh above the ground, \emph{Penetration} (Pen) the average depth by which the mesh sinks below it, and \emph{Foot-contact} their sum. Writing $z_i$ for the signed height of the lowest mesh vertex in frame $i$,
\begin{equation}
    \mathrm{Float}=\frac{1}{N}\sum_{i=1}^{N}\max(0,z_i),
    \quad
    \mathrm{Pen}=\frac{1}{N}\sum_{i=1}^{N}\max(0,-z_i),
\end{equation}
\noindent and $\text{Foot-contact}=\mathrm{Float}+\mathrm{Pen}$. Hovering and penetration are opposite artefacts, so it is their sum, rather than either half alone, that certifies contact realism.

\noindent \textbf{Why this set represents physical fidelity.} FID measures the distributional divergence between generated and real motions, i.e.\ how closely the generation follows real human-motion statistics as a whole, and is thus a \emph{macro-level} statement about physical plausibility. It is, however, largely blind to local violations: mild foot sliding, a few centimetres of hovering or brief ground penetration still fall well inside the real-motion distribution. This is why the macro level alone cannot certify physical fidelity, and why we pair it with the three \emph{micro-level} metrics marked as such in Table~\ref{tab:metric_property}: PFC certifies dynamics feasibility (acceleration must be explained by foot--ground contact), while Float and Foot-contact certify contact realism (the body neither hovers above the floor nor sinks into it). Macro and micro together are what we mean by physical fidelity: the first keeps the generated distribution close to that of real motion as a whole, the second checks that each individual sequence respects the ground it stands on.

\input{Tables/table_metric_property}

\section{User Study}
\label{app:user}

We conduct rigorous user studies to evaluate both motion quality and semantic alignment. All tests employ \textit{pairwise} comparisons between our method and six baselines: GT, MDM, MoMask, T2M-GPT, ParCo, and BAMM. For each comparison, we generate 40 motions using identical text prompts from the HumanML3D test set.

\noindent\textbf{Protocol.} Participants view two video clips synthesized by different methods and select their preferred one based on two criteria: (1) \emph{``Which motion more closely resembles that of a real human?''} for motion quality, and (2) \emph{``Which motion more aligns with the given text?''} for semantic alignment. Before testing, each participant completes a training page with fixed example videos to ensure task familiarity. During testing, attention checks are randomly inserted—displaying the message \emph{``Attention: Please select the right motion''}—and responses failing these checks are excluded from analysis.

\noindent\textbf{Results.} We recruit 38 participants with qualified English proficiency, each receiving 6 GBP compensation (average completion time: 30 minutes). All participants pass the filtering criteria. As shown in Figures~\ref{fig:img1} and~\ref{fig:img2}, our method consistently outperforms all competitors in both motion naturalness and semantic accuracy. Notably, our approach achieves approximately 40\% preference rate even against ground-truth samples, demonstrating competitive human-like motion quality.
\input{NewFigures/fig_user_study}
\section{Additional Ablation Studies}
\label{app:ablate}

\noindent\textbf{Impact of Classifier-Free Guidance.}
We analyze the sensitivity of our LMR framework to the guidance scale $s$ during inference (Figure~\ref{fig:cfg_impact}). For the \textit{Continuous Backbone}, performance peaks at $s{=}5$, achieving FID of \textbf{9.937} and Top-1 R-Precision of \textbf{0.648}, suggesting that continuous latent spaces require stronger guidance to sharpen the probability density towards the text condition. For the \textit{Discrete Backbone}, the optimal scale is $s{=}2$ with FID of \textbf{0.038} and Top-1 R-Precision of \textbf{0.526}. Increasing $s$ beyond this point causes rapid quality degradation, indicating that excessive guidance in discrete spaces leads to probability mass collapse.
\input{NewFigures/Fig_cfg_scale_impact}

\input{Tables/table_Efficiency}

\noindent\textbf{ Efficiency Comparison (Table~\ref{tab:efficiency}).}
Table~\ref{tab:efficiency} compares training and inference efficiency against T2M-GPT, ParCo, and MoMask.
Compared to T2M-GPT, our method introduces additional reasoning layers in the tokenizer, resulting in slightly more parameters and higher per-frame latency. However, this overhead yields significantly better generation quality.
ParCo and MoMask reduce network capacity to accelerate inference, but ParCo requires six tokenizer components leading to longer training time, while MoMask needs six sequential passes for RVQ decoding. Notably, neither supports KV caching. In contrast, our autoregressive generator enables KV caching~\cite{KVcache}, reducing inference time to 0.4 ms/frame.

\section{Hyper-parameter Sensitivity}
\label{app:hparam}
We complement the token-configuration ablation in the main text (Table~\ref{tab:ablation_settings}, which already covers the reasoning downsampling rate) with a sensitivity study of two further key hyper-parameters in Table~\ref{tab:hparam_sensitivity}: the relative weighting of the two reasoning-branch losses $\lambda_{align}\!:\!\lambda_{mtp}$, and the reasoning codebook size. Across both, LMR varies smoothly with no cliff-like degradation. The retrieval and distance metrics (Top-1, MM-Dist) are particularly robust, staying within a narrow band across all settings. FID shows a comparatively larger swing, but in a \emph{regular, well-behaved} manner: it changes monotonically and unimodally around the optimum (a clear single ``valley'' at $0.5\!:\!0.1$ and at codebook size $512$), so it is easy to tune and the optimum is unambiguous rather than a brittle, hard-to-locate sweet spot. Overall, this confirms that our design choices are robust rather than brittle.

\input{Tables/table_hparam_sensitivity}

\section{Pseudo-code of LMR}
\label{app:pseudocode}
For reproducibility, we summarize the tokenizer training, generator training, and inference in Algorithms~\ref{alg:tok}--\ref{alg:inf} (discrete Scenario~A, with the T2M-GPT backbone). Consistent with our stepwise training, the execution branch ($\mathcal{E}, \mathcal{Z}_{exec}, \mathrm{Dec}$) is first trained as a plain motion VQ and then frozen; only the reasoning head is trained on top. The continuous Scenario~B is structurally identical, replacing VQ with a KL-regularized continuous latent and per-token softmax sampling with per-token diffusion denoising.

\begin{algorithm}[h]
\caption{Dual-Granularity Tokenizer Training (stepwise)}
\label{alg:tok}
\begin{algorithmic}[1]
\STATE \textbf{Input:} motion batch $\mathbf{x}$; text prompt $\mathbf{w}$
\STATE \textbf{Frozen (Stage~0):} encoder $\mathcal{E}$, execution codebook $\mathcal{Z}_{exec}$, decoder $\mathrm{Dec}$
\STATE \textbf{Trained:} projector $\Phi_{res}$, reasoning codebook $\mathcal{Z}_{res}$, MTP decoder $\mathrm{Dec}_{mtp}$
\STATE $\mathbf{f} \leftarrow \mathcal{E}(\mathbf{x})$ \quad // frozen encoder
\STATE $\mathbf{q}_{exec} \leftarrow \arg\min_k \|\mathbf{f}-\mathcal{Z}_{exec}^{(k)}\|_2$ \quad // length $T$, frozen
\STATE $\mathbf{f}_{res} \leftarrow \Phi_{res}(\mathbf{f})$ \quad // project + downsample to $T/4$
\STATE $\mathbf{q}_{res} \leftarrow \arg\min_k \|\mathbf{f}_{res}-\mathcal{Z}_{res}^{(k)}\|_2$;\quad $\hat{\mathbf{f}}_{res}\leftarrow \mathcal{Z}_{res}(\mathbf{q}_{res})$
\STATE $\mathcal{L}_{align} \leftarrow 1-\cos(\mathrm{AvgPool}(\hat{\mathbf{f}}_{res}), \mathrm{BERT}(\mathbf{w}))$
\STATE $\mathcal{L}_{mtp} \leftarrow \mathrm{CE}(\mathrm{Dec}_{mtp}(\hat{\mathbf{f}}_{res}), \mathbf{w}_{masked})$
\STATE $\mathcal{L} \leftarrow \lambda_{align}\mathcal{L}_{align}+\lambda_{mtp}\mathcal{L}_{mtp}+\mathcal{L}_{VQ}(\mathbf{f}_{res},\hat{\mathbf{f}}_{res})$
\STATE update only $\{\Phi_{res}, \mathcal{Z}_{res}, \mathrm{Dec}_{mtp}\}$ \quad // $\mathcal{E},\mathcal{Z}_{exec},\mathrm{Dec}$ frozen
\STATE \textbf{Output:} frozen tokenizer $(\mathcal{E}, \mathcal{Z}_{res}, \mathcal{Z}_{exec}, \mathrm{Dec})$
\end{algorithmic}
\end{algorithm}

\begin{algorithm}[h]
\caption{Generator Training (Unified Autoregressive)}
\label{alg:gen}
\begin{algorithmic}[1]
\STATE \textbf{Input:} motion $\mathbf{x}$, text $c$; frozen tokenizer; CLIP encoder
\STATE \textbf{Components:} causal Transformer $\mathcal{H}_\theta$; dual heads $\mathbf{W}_{res}, \mathbf{W}_{exec}$
\STATE $\mathbf{q}_{res}, \mathbf{q}_{exec} \leftarrow \mathrm{Tokenizer}(\mathbf{x})$;\quad $S \leftarrow [\mathbf{q}_{res}; \mathbf{q}_{exec}]$
\STATE $e_c \leftarrow \mathrm{CLIP}(c)$; with prob.\ $p_{uncond}$ set $e_c\leftarrow\emptyset$ \quad // CFG
\FOR{$t = 1$ \TO $|S|$}
  \STATE $h_t \leftarrow \mathcal{H}_\theta(S_{<t}, e_c)$
  \IF{$t \le T/4$} \STATE $\text{logits}_t \leftarrow \mathbf{W}_{res}^{\top} h_t$ \quad // reasoning head
  \ELSE \STATE $\text{logits}_t \leftarrow \mathbf{W}_{exec}^{\top} h_t$ \quad // execution head
  \ENDIF
\ENDFOR
\STATE $\mathcal{L}_{ce} \leftarrow \sum_t \mathrm{CE}(\text{logits}_t, S_t)$; update $\{\mathcal{H}_\theta, \mathbf{W}_{res}, \mathbf{W}_{exec}\}$
\STATE \textbf{Output:} trained generator $\mathcal{H}_\theta$
\end{algorithmic}
\end{algorithm}

\begin{algorithm}[h]
\caption{Inference (``Think-then-Act'')}
\label{alg:inf}
\begin{algorithmic}[1]
\STATE \textbf{Input:} text $c$; frozen $\mathcal{H}_\theta$; decoder $\mathrm{Dec}$; guidance scale $s$, top-$p$
\STATE $S \leftarrow [\,]$;\quad $e_c \leftarrow \mathrm{CLIP}(c)$
\STATE \textit{// Phase~I: Reasoning (plan)}
\FOR{$t = 1$ \TO $T/4$}
  \STATE $\text{logits} \leftarrow (1{+}s)\,\mathbf{W}_{res}^{\top}\mathcal{H}_\theta(S, e_c) - s\,\mathbf{W}_{res}^{\top}\mathcal{H}_\theta(S, \emptyset)$
  \STATE $S_t \sim \mathrm{top\text{-}}p(\mathrm{softmax}(\text{logits}))$;\quad $S.\mathrm{append}(S_t)$
\ENDFOR
\STATE \textit{// Phase~II: Execution (act)}
\FOR{$t = T/4{+}1$ \TO $T/4{+}T$}
  \STATE $\text{logits} \leftarrow (1{+}s)\,\mathbf{W}_{exec}^{\top}\mathcal{H}_\theta(S, e_c) - s\,\mathbf{W}_{exec}^{\top}\mathcal{H}_\theta(S, \emptyset)$
  \STATE $S_t \sim \mathrm{top\text{-}}p(\mathrm{softmax}(\text{logits}))$;\quad $S.\mathrm{append}(S_t)$
\ENDFOR
\STATE $\mathbf{q}_{exec} \leftarrow S[T/4{+}1:]$;\quad $\hat{\mathbf{x}} \leftarrow \mathrm{Dec}(\mathcal{Z}_{exec}(\mathbf{q}_{exec}))$
\STATE \textbf{Output:} generated motion $\hat{\mathbf{x}}$
\end{algorithmic}
\end{algorithm}

%% file: Tables/table_metric_property.tex
\begin{table*}[t]
\centering
\caption{Full name, informal definition and certified property of every metric reported in this paper. Physical fidelity is a two-level concept: FID states it at the \emph{macro} (distributional) level, while PFC, Float and Foot-contact state it at the \emph{micro} (per-sequence, per-frame) level. Retrieval metrics measure the separate property of semantic alignment, and MPJPE measures reconstruction fidelity.}
\label{tab:metric_property}
\setlength{\tabcolsep}{5pt}
\renewcommand{\arraystretch}{1.15}
\resizebox{\textwidth}{!}{
\begin{tabular}{llll}
\toprule
Abbrev. & Full name & Definition (informal) & Certified property \\
\midrule
FID & Fr\'echet Inception Distance
    & Fr\'echet distance between real / generated feature Gaussians
    & Physical fidelity (\emph{macro}) \\
rFID & reconstruction FID
    & FID computed on tokenizer reconstructions
    & Reconstruction fidelity \\
gFID & generation FID
    & FID computed on text-conditioned generations
    & Physical fidelity (\emph{macro}) \\
\midrule
PFC & Physical Foot Contact score~\cite{EDGE}
    & Centre-of-mass acceleration co-occurring with both feet in motion
    & Physical fidelity (\emph{micro}) \\
Float & Float~\cite{PhysiInter}
    & Average height of the lowest point of the human mesh above the ground
    & Physical fidelity (\emph{micro}) \\
Foot-contact & Foot Contact~\cite{PhysiInter}
    & Float $+$ Penetration: total departure from exact ground contact
    & Physical fidelity (\emph{micro}) \\
\midrule
R-Precision & Retrieval Precision (Top-1/2/3)
    & Top-$k$ retrieval accuracy against 1 true and 31 mismatched texts
    & Semantic alignment \\
MM-Dist & Multimodal Distance
    & Mean distance between paired motion and text features
    & Semantic alignment \\
\midrule
Diversity & Diversity
    & Mean pairwise distance over randomly sampled motion pairs
    & Distributional spread ($\rightarrow$) \\
MPJPE & Mean Per Joint Position Error
    & Root-aligned per-joint position error (mm)
    & Reconstruction fidelity \\
ACC & masked-keyword prediction Accuracy
    & Accuracy of recovering masked keywords from the reasoning tokens
    & Local reasoning alignment \\
\bottomrule
\end{tabular}
}
\end{table*}

%% file: NewFigures/fig_user_study.tex
\begin{figure}[htbp]
\centering
\begin{minipage}{.25\textwidth}
  \centering
  \includegraphics[width=0.9\linewidth]{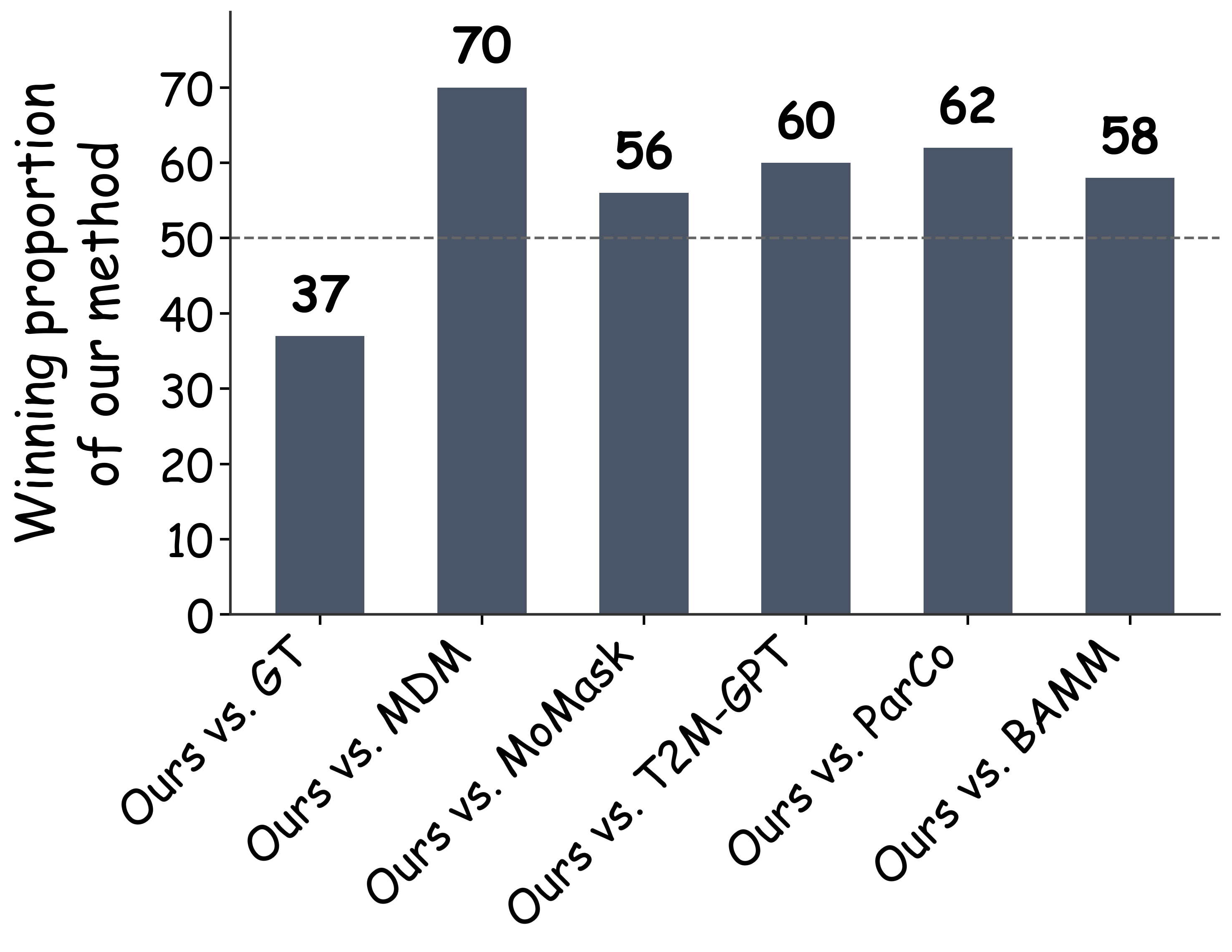}
  \caption{User study on motion quality.}
  \label{fig:img1}
\end{minipage}%
\begin{minipage}{.25\textwidth}
  \centering
  \includegraphics[width=0.9\linewidth]{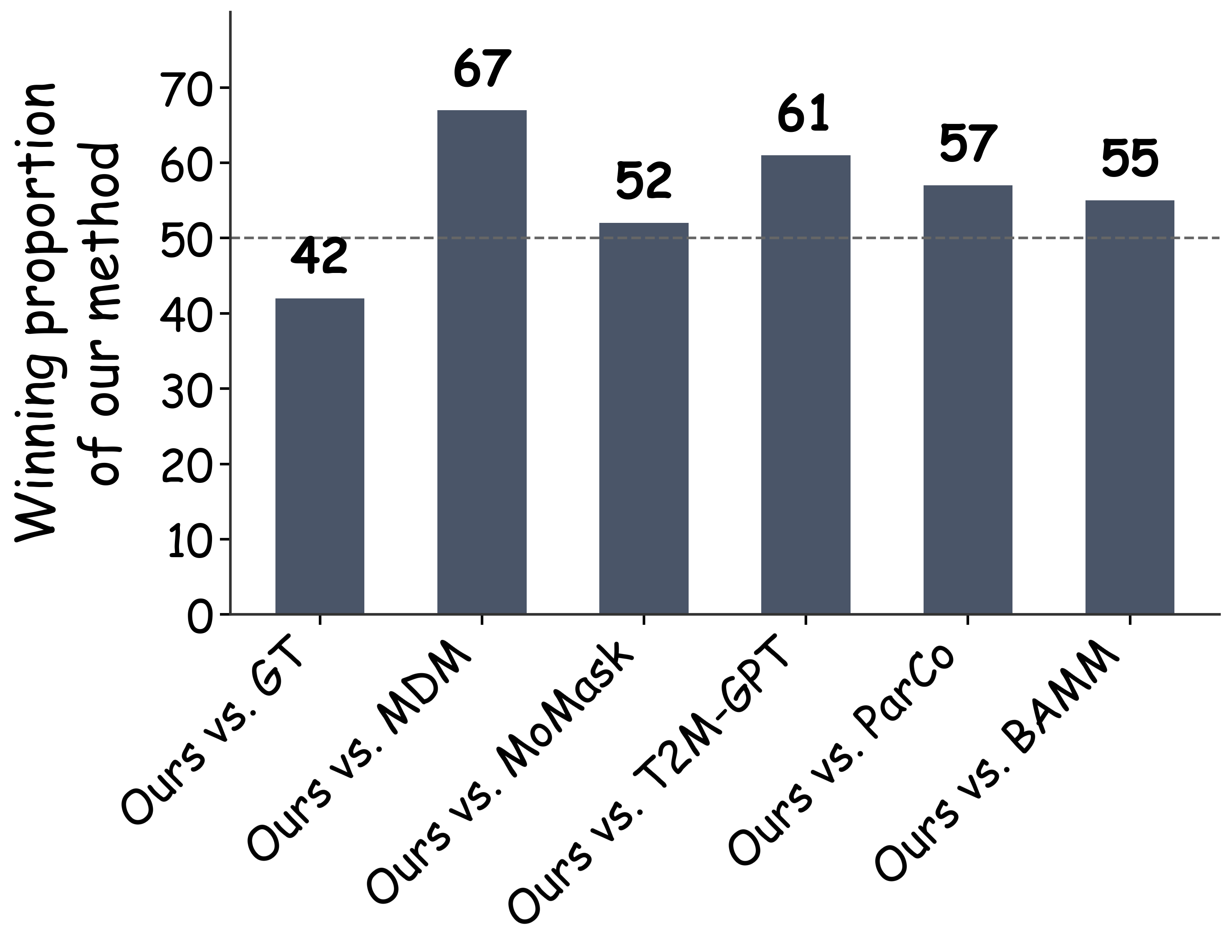}
  \caption{User study on semantics.}
  \label{fig:img2}
\end{minipage}
\end{figure}

%% file: NewFigures/Fig_cfg_scale_impact.tex
\begin{figure}[t]
    \centering
    \includegraphics[width=\linewidth]{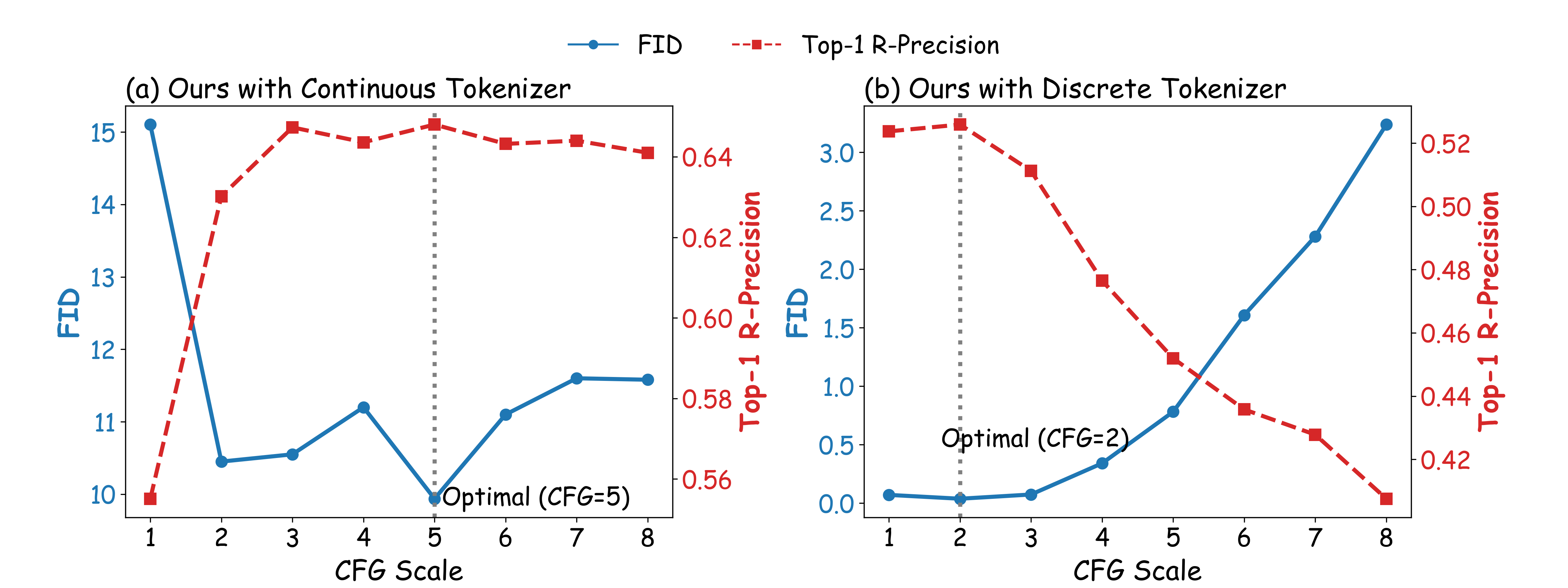}
    \caption{Impact of Classifier-Free Guidance scale on generation quality and semantic alignment. We evaluate FID and Top-1 R-Precision across varying scales. The continuous MotionStreamer backbone requires a higher guidance scale of 5 for optimal performance. Conversely, the discrete T2M-GPT backbone peaks at a lower scale of 2, suggesting that discrete representations are more sensitive to guidance intensity.}
    \label{fig:cfg_impact}
\end{figure}

%% file: Tables/table_Efficiency.tex
\begin{table}[htbp]
    \centering
    \caption{Efficiency evaluation during training and inference. * employs KV caching to speed up autoregressive generation. Tok. and Gen. denote Tokenizer and Generator.}
    \label{tab:efficiency}
    \resizebox{\columnwidth}{!}{%
    \begin{tabular}{lccccccc}
        \toprule
        \multirow{2}{*}{\textbf{Method}} & \multicolumn{2}{c}{\textbf{Infer.} (ms/frame)} & \multicolumn{2}{c}{\textbf{Train.} (H)} & \multicolumn{2}{c}{\textbf{Size} (M)} \\
        \cmidrule(lr){2-3} \cmidrule(lr){4-5} \cmidrule(lr){6-7}
         & Tok. & Gen. & Tok. & Gen. & Tok. & Gen. \\
        \midrule
        T2M-GPT~\cite{T2M-GPT}  & 1.1 & 31 & 15 & 40 & 19.44 & 228.42 \\
        ParCo~\cite{ParCo}  & 1.2 & 5 & 4 & 18 & 6.35 & 19.44 \\
        MoMask~\cite{MoMask}  & 2.8 & 15 & 18 & 4 & 25.41 & 166.05 \\
        \midrule
        \textbf{Ours} & 2.4 & 30 & 5 & 19 & 24.79 & 229.62 \\
        \textbf{Ours}* & - & \textbf{0.4} & - & - & - & - \\
        \bottomrule
    \end{tabular}%
    }
\end{table}

%% file: Tables/table_hparam_sensitivity.tex
\begin{table}[h]
\centering
\caption{Sensitivity of LMR to two key hyper-parameters on HumanML3D. Since the reasoning branch is trained in a separate stage whose only objectives are $\mathcal{L}_{align}$ and $\mathcal{L}_{mtp}$, the first study only varies their relative weighting $\lambda_{align}\!:\!\lambda_{mtp}$ (with $\lambda_{align}$ fixed at $0.5$); the second varies the reasoning codebook size. Each point retrains the tokenizer reasoning head and the generator. Performance varies smoothly with no cliff-like degradation. $^{\ast}$ denotes the main configuration.}
\label{tab:hparam_sensitivity}
\setlength{\tabcolsep}{4pt}
\resizebox{\columnwidth}{!}{
\begin{tabular}{llccc}
\toprule
Hyper-parameter & Setting & FID $\downarrow$ & Top1 $\uparrow$ & MM-Dist $\downarrow$ \\
\midrule
\multirow{3}{*}{$\lambda_{align}\!:\!\lambda_{mtp}$}
 & $0.5\!:\!0.05$        & 0.088 & 0.511 & 3.048 \\
 & $0.5\!:\!0.1^{\ast}$  & \textbf{0.040} & \textbf{0.537} & \textbf{2.895} \\
 & $0.5\!:\!0.2$         & 0.075 & 0.505 & 3.063 \\
\midrule
\multirow{3}{*}{Reasoning codebook}
 & 256          & 0.072 & 0.510 & 3.047 \\
 & 512$^{\ast}$ & \textbf{0.040} & \textbf{0.537} & \textbf{2.895} \\
 & 1024         & 0.071 & 0.518 & 3.012 \\
\bottomrule
\end{tabular}
}
\end{table}